\documentclass[10pt,twocolumn,letterpaper]{article} 

\usepackage{avss}
\usepackage{times}
\usepackage{epsfig}
\usepackage{graphicx}
\usepackage{amsmath}
\usepackage{amssymb}
\usepackage{subcaption}
\usepackage{color}
\usepackage[multiple]{footmisc}
\avssfinalcopy % *** Uncomment this line for the final submission

\def\avssPaperID{90} % *** Enter the AVSS Paper ID here

\begin{document}

%%%%%%%%% TITLE
%\title{Video-based Bottleneck detection based on Crowd Movement }
%\title{Lagrangian-based Bottleneck Detection for Video-Surveillance}
%\title{Lagrangian-based Bottleneck Detection}
%\title{Video-based Bottleneck detection a Lagrangian approach}
%\title{Video-based Bottleneck detection a Lagrangian perspective}
%\title{Video-based Bottleneck detection utilizing Lagrangian fields}
\title{Video-based Bottleneck Detection utilizing Lagrangian Dynamics in Crowded Scenes}
%\title{Video-based Bottleneck detection utilizing FTLE}
%\title{Where are we going to? Video-based Bottleneck detection utilizing Lagrangian Dynamics}
% \title{Getting sucked/fucked in a Crowd: Video-based Bottleneck detection utilizing Lagrangian Dynamics}
% \title{Getting stuck in a Crowd: Video-based Bottleneck detection utilizing Lagrangian Dynamics}
% \title{Getting lost in the Bottle(neck): Video-based Bottleneck detection utilizing Lagrangian Dynamics}

\author{Maik Simon, Markus K\"uchhold, Tobias Senst, Erik Bochinski, Thomas Sikora\\
Communication Systems Group\\
Technische Universit\"at Berlin\\
{\tt\small simon, kuechhold, senst, bochinski, sikora@nue.tu-berlin.de}
% \and
% {\tt\small secondauthor@i1.org}
}

%% $Id: macros.tex,v 1.6 2005/04/21 03:46:23 bzfweink Exp $
%% Based on a TeXnicCenter-Template by Tino Weinkauf.
%%%%%%%%%%%%%%%%%%%%%%%%%%%%%%%%%%%%%%%%%%%%%%%%%%%%%%%%%%%%%

%%%%%%%%%%%%%%%%%%%%%%%%%%%%%%%%%%%%%%%%%%%%%%%%%%%%%%%%%%%%%
%% OPTIONS
%%%%%%%%%%%%%%%%%%%%%%%%%%%%%%%%%%%%%%%%%%%%%%%%%%%%%%%%%%%%%

\newcommand{\methodname}{{TODO }}
\newcommand{\RRR}{{\mathrm I\! \mbox{R} }}
\newcommand{\EEE}{{\mathrm I\! \mbox{E} }}
\newcommand{\xx}{{\mathbf x}}
\newcommand{\yy}{{\mathbf y}}
\newcommand{\zz}{{\mathbf z}}
\newcommand{\dd}{{\mathbf d}}
\newcommand{\hh}{{\mathbf h}}
\newcommand{\ttt}{{\mathbf t}}
\newcommand{\jj}{{\mathbf j}}
\newcommand{\ii}{{\mathbf i}}
\newcommand{\ff}{{\mathbf f}}
\newcommand{\pp}{{\mathbf p}}
\newcommand{\qq}{{\mathbf q}}
\newcommand{\aaa}{{\mathbf a}}
\newcommand{\ssss}{{\mathbf s}}
\newcommand{\bb}{{\mathbf b}}
\newcommand{\ee}{{\mathbf e}}
\newcommand{\cc}{{\mathbf c}}
\newcommand{\hc}{{\hat{c}}}
\newcommand{\nn}{{\mathbf n}}
\newcommand{\mm}{{\mathbf m}}
\newcommand{\kk}{{\mathbf k}}
\newcommand{\rr}{{\mathbf r}}
\newcommand{\vv}{{\mathbf v}}
\newcommand{\AAA}{{\mathbf A}}
\newcommand{\JJ}{{\mathbf J}}
\newcommand{\CC}{{\mathbf C}}
\newcommand{\DD}{{\mathbf D}}
\newcommand{\BB}{{\mathbf B}}
\newcommand{\EE}{{\mathbf E}}
\newcommand{\PP}{{\mathbf P}}
\newcommand{\MM}{{\mathbf M}}
\newcommand{\QQ}{{\mathbf Q}}
\newcommand{\FF}{{\mathbf F}}
\newcommand{\GG}{{\mathbf G}}
\newcommand{\TT}{{\mathbf T}}
\newcommand{\RR}{{\mathbf R}}
\newcommand{\SSS}{{\mathbf S}}
\newcommand{\Sa}{{\mathbf {Sa}}}
\newcommand{\Topo}{{\mathbf {Topo}}}
\newcommand{\Sep}{{\mathbf {Sep}}}
\newcommand{\Str}{{\mathbf {Str}}}
\newcommand{\Reg}{{\mathbf {Reg}}}
\newcommand{\cs}{{\dot c}}
\newcommand{\sss}{{\dot s}}
\newcommand{\ps}{{\dot p}}
\newcommand{\pg}{{\dot g}}
\newcommand{\ww}{{\mathbf w}}
\newcommand{\xxs}{\dot{\mathbf x}}
\newcommand{\vvs}{\dot{\mathbf v}}
\newcommand{\us}{{\dot u}}
\newcommand{\ws}{{\dot w}}
\newcommand{\fn}{{\mathbf 0}}
\newcommand{\nv}{{\mathbf 0}}
\newcommand{\sssd}{{\dot {\mathbf s}}}
\newcommand{\ZV}{{\mathbf 0}}
\newcommand{\D}{d}
\newcommand{\bsigma}{{\mbox{\boldmath$\sigma$}}}
\newcommand{\bdelta}{{\mbox{\boldmath$\delta$}}}
% definition of a feature point
\newcommand{\fp}{{\mathbf p}}

\newcommand{\btheta}{\theta}
\newcommand{\residual}{{r_i(\btheta)}}
\newcommand{\sO}{{s_{\Omega}}}

\newcommand{\li}{\item}

\newcommand{\vecin}{\left(\begin{array}{c}}
\newcommand{\vecout}{\end{array}\right)}
\newcommand{\lbr}{\left\{}
\newcommand{\rbr}{\right\}}
\newcommand{\sk}{s}
\newcommand{\sbf}{\mathbf{s}}
\newcommand{\cluster}{\mathbf{c}}
\newcommand{\feature}{\mathbf{f}}
\newcommand{\sfat}{\small \textbf}
\newcommand{\tfat}{\tiny \textbf}
\newcommand{\arcL}{$\Lambda_{arcL}$ }

\newcommand{\F}[1]{\textbf{#1} } %% for equations
\newcommand{\refEq}[1]{Eq.~(\ref{#1})} %% for equations
\newcommand{\refeq}[1]{Eq.~(\ref{#1})} %% for equations in the text
\newcommand{\refFig}[1]{Figure~\ref{#1}}
\newcommand{\reffig}[1]{Figure~\ref{#1}}   %% for figures in the text
\newcommand{\refTab}[1]{Table~\ref{#1}}   %% for table in the 
\newcommand{\reftab}[1]{Table~\ref{#1}}   %% for figures in the text
\newcommand{\refSec}[1]{Section~\ref{#1}}   %% for figures in the text
\newcommand{\refsec}[1]{Section~\ref{#1}}   %% for figures in the text
\newcommand{\refchap}[1]{Chapter~\ref{#1}}   %% for figures in the text
\newcommand{\refChap}[1]{Chapter~\ref{#1}}   %% for figures in the text
\newcommand{\refapp}[1]{Appendix~\ref{#1}}   %% for figures in the text
\newcommand{\reftheo}[1]{Theorem~\ref{#1}}   %% for figures in the text
\newcommand{\refTheo}[1]{Theorem~\ref{#1}}   %% for figures in the text

\newcommand{\FormatSequence}[1]{#1}   %% entscheid wie das format einer sequenc (ausgeschrieben) aussieht
\newcommand{\AEERR}{AEE & tAEE }
\newcommand{\AEER}{AEE & R3 }
\newcommand{\GroundThruth}{ground-truth }   %% entscheid wie das format einer sequenc
\newcommand{\groundthruth}{ground-truth }   %% entscheid wie das format einer
\newcommand{\groundthruthdot}{ground-truth. }  
\newcommand{\pathline}{path-line }   %% entscheid wie das format einer sequenc
\newcommand{\pathlines}{path-lines }   %% entscheid wie das format einer sequenc
\newcommand{\pathlinesdot}{path-lines. }   %% entscheid wie das format einer sequenc
\newcommand{\pathlinescomma}{path-lines, }   %% entscheid wie das format einer sequenc
\newcommand{\pathlinedot}{path-line. }   %% entscheid wie das format einer sequenc
\newcommand{\Pathlines}{Path-lines }   %% entscheid wie das format einer sequenc
\newcommand{\Pathline}{Path-lines }   %% entscheid wie das format einer sequenc
\newcommand{\Streamlines}{Stream-lines }
\newcommand{\streamlines}{stream-lines }
\newcommand{\streamlinescomma}{stream-lines, }
\newcommand{\streamline}{stream-line }
\newcommand{\streamlinecomma}{stream-line, }
\newcommand{\streamlinesdot}{stream-lines. }
\newcommand{\Streaklines}{Streak-lines }
\newcommand{\streaklines}{streak-lines }
\newcommand{\streakline}{streak-line }
\newcommand{\Timelines}{Time-lines }

\newcommand{\Sintel}{MPI-Sintel }   %% entscheid wie das format einer sequenc
\newcommand{\runtime}{run-time }
\newcommand{\runtimedd}{run-time: }
\newcommand{\runtimecomma}{run-time, }
\newcommand{\runtimedot}{run-time. }
\newcommand{\runtimes}{run-times }
\newcommand{\Runtimes}{Run-times }
\newcommand{\realtime}{real-time }
\newcommand{\realtimedot}{real-time. }
\newcommand{\realtimecomma}{real-time, }
\newcommand{\publicationTime}{28.10.2016 }
 %(ausgeschrieben) aussieht
\newcommand{\dataset}{dataset }
\newcommand{\datasetdot}{dataset. }
\newcommand{\datasetdd}{dataset: }
\newcommand{\datasets}{datasets }
\newcommand{\datasetsdot}{datasets. }
\newcommand{\datasetscomma}{datasets, }
\newcommand{\datasetcomma}{dataset, }
\newcommand{\Endpoint}{end-point }
\newcommand{\Endpoints}{end-points }
\newcommand{\tFTLE}{\overline{\text{FTLE}}}

%%%
%%% Style Definitions
%%%
\newcommand{\texttodo}[1]{\noindent\textit{\textcolor{blue}{TODO: #1}}}

\newcommand{\vb}{ \left( \begin {array} {c} }
\newcommand{\ve}{ \end {array} \right) }
\newcommand{\matin}{ \begin {bmatrix} }
\newcommand{\matout}{ \end {bmatrix} }
\newcommand{\surfimgwidth}{0.3\linewidth}
\newcommand{\sketchimgheight}{10em}
\newcommand{\eqinitspace}{7em}
\newcommand{\leftindent}{\hspace{-90pt}}

\newcommand{\SalsaWidth}{ 0.37\textwidth}
\renewcommand{\topfraction}{0.75}
\renewcommand{\textfraction}{0.1}
\renewcommand{\floatpagefraction}{0.75}
\renewcommand{\emph}[1]{\textit{#1}}

\newcommand{\emptypage}
{
	\thispagestyle{empty}
	\mbox{}
	\newpage
}
		
\newcommand{\chappage}[2]{
    \thispagestyle{empty}
    {
        \centering
        {\color{rule_color}\textsc{Doctoral Thesis}}\\[30em]
        \textrm{Chapter} #1\\[2em]
        \color{Maroon}\textbf{\Large\textsc{#2}}\\[1.5em]
        {\color{rule_color}\titlerule}
    }
    \newpage
}

\newcommand{\borderheading}[1]
{
    \vspace{0.5em}
    \graffito{\vfill\vspace{2em}\textbf{#1}}
}

\maketitle
% \thispagestyle{empty}
%
%%%%%%%%% ABSTRACT
\begin{abstract}
Avoiding bottleneck situations in crowds is critical for the safety and comfort of people at large events or in public transportation.
Based on the work of Lagrangian motion analysis we propose a novel video-based bottleneck-detector by identifying characteristic stowage patterns in crowd-movements captured by optical flow fields.
%These movements are regarded as a flowing continuum which is captured by optical flow.
The Lagrangian framework allows to assess complex time-dependent crowd-motion dynamics at large temporal scales near the bottleneck by two dimensional Lagrangian fields.
In particular we propose long-term temporal filtered Finite Time Lyapunov Exponents (FTLE) fields that provide towards a more global segmentation of the crowd movements and allows to capture its deformations when a crowd is passing a bottleneck.
Finally, these deformations are used for an automatic spatio-temporal detection of such situations.
% \texttodo{The performance of the proposed approach is shown in extensive evaluations on the existing J\"ulich and AGORASET  datasets,  that  we  have  updated  with  ground truth data for spatio-temporal bottleneck analysis. Furthermore a metric is introduced.} 
The performance of the proposed approach is shown in extensive evaluations on the existing J\"ulich and AGORASET  datasets,  that  we  have  updated  with  ground truth data for spatio-temporal bottleneck analysis.
%\texttodo{Furthermore, the existing AGORASET and Jülich data sets were updated with Ground Truth data for recognition purposes. A metric is presented for the evaluation of the data}
%letzter satz muss besser werden, der passt in 90% aller paper.
% \newline
% \textbf{Keywords-} crowd analysis; Lagrangian framework; FTLE; bottleneck detection;
\end{abstract}

%%%%%%%%% BODY TEXT

%introduction

\section{Introduction}

The analysis of crowd movements is of importance for the safety and comfort of people in transport infrastructures. 
Handling crowded scenes during public events (e.g.$ $ fan parks, concerts, sport events) is a challenging task for security personnel, police and crisis management teams. 
Especially the occurrence of bottlenecks during an event can lead to panics due to overcrowding.
An automatic bottleneck identification system can aid the operator to prevent critical situations by assessing characteristic crowd-movement patterns.
The aim of this work is to identify such events in the spatial and temporal domain.

%One major area of research is the development of intelligent systems that allow automated event recognition in crowded scenes.
%In this paper we want to detect bottlenecks based on visual surveillance by analyzing the movement of pedestrians in crowded scenes.
%
In computer vision the analysis of high density crowds is performed on macroscopic perspective \cite{Silveira_Jacques_jr_2010}, i.e.$ $ the crowd is assessed as a single entity.
The behaviors of individuals in a crowd are dependent on the crowd behavior \cite{Silveira_Jacques_jr_2010,Li_2015} and modelled by fluid dynamic processes \cite{Ali_2007, Wu2010,Moore2011, Kuhn_2012}.
Hughes work \cite{hughes_2003} supports the assumption that crowds are a flowing continuum and proposed three main behavioral hypotheses for persons moving in a crowd.
In \cite{Bain46}, Bain and Bartolo also contemplate pedestrian flows with the help of a hydrodynamic model. % too. %
Here, the flow behavior of polarized crowds was examined by considering the border movements of the crowd at the start of various marathons. 
%
%
%For the analysis of the behaviour of crowds there are manifold studies.
% Tobi: \textbf{Vielleicht weg lassens ->} %find ich an sich gut für den einstieg, EB
To describe crowd behavior for crowd simulation Still proposed in \cite{Still_2000} three main effects: 
% Still \cite{Still_2000} describes three major types of crowd behavior:
i.) least-effort hypothesis means that people are looking for the least strenuous route 
ii.) lane formation implies that people walk most easily behind each other and
iii.) bottleneck effect occurs at a narrowing point with a significant speed change of the crowd and represents at the same time a critical point. 
%
%\textbf{<-}
%
% seyfried_2009 sind dann wiederum 12 Videos von 80 aus dem verwendeten datensatz
%Several researches show that the behaviour before and within a bottleneck is influenced by different aspects.

The following studies investigate the influence of the bottleneck to the behaviour of the crowd before and within the narrow pass.
Seyfried \etal \cite{seyfried_2009} shows an experimental study in which the flow of unidirectional pedestrian streams through bottlenecks was evaluated.
The result was a linear growth of the flow with a simultaneous increase in the width of the bottleneck and the observation of the phenomenon of lane formation within bottlenecks.
%For this purpose, individual velocities, local densities and time gaps for different bottleneck widths were investigated. 
%This showed a linear growth of the flow with a simultaneous increase in the width of the bottleneck. The work  engages with the phenomenon of lane formation within bottlenecks.
%
%
%kruechten_2016 ist die passende Quelle zu 60 Videos des verwendeten Datensatzes..
Kr\"uchten \etal \cite{kruechten_2016} recorded a dataset under laboratory conditions, which represents persons with different age, group sizes and social group sizes in case of evacuation through a bottleneck. %TODO Hallo liest du noch? 2x group sizes? --> einmal ist es die gesamtgruppen Größe die durch das Bottleneck gelotst wird und es gibt dann wiederum nochmal die sozialen Gruppen innerhalb der Gesamtgruppe. Sprich einmal 50 Leute durchlaufen.. alles Einzelkämpfer und 50 Leute wobei 2 jeweils ein Paar bilden
In the study, the social aspect of passing through a bottleneck was presented, which showed that with increasing social group strength, the flow through the bottleneck increased.
% sieben_2017 --> entrance sequenzen
The study of Sieben \etal \cite{sieben_2017}, showed the influence of the spatial structure and the perception of the participants in comparison to physical measurements.

The data recorded in \cite{Liao_2014,seyfried_2009,sieben_2017,kruechten_2016} has been published at the pedestrian dynamic archive and will be denoted as J\"ulich dataset. % set.
Allain \etal \cite{allain_2012} proposed the AGORASET for crowd behaviour analysis containing corridors, obstacles and escapes.
The dataset consists of synthetic rendered images and provides a higher variation and different point of views of the scenes.

Each scene of the two datasets contains physical bottlenecks which are not always leading to congestion situations in pedestrian movements.
Bottleneck situations can also arise through situation-dependent events, whereby an occuring bottleneck is defined by the flow of movement of the persons.
For this reason, the datasets were extended with ground truth data that has both spatial and temporal properties. 
Furthermore a new metric is presented for this spatio-temporal problem, which makes the proposed and future methods comparable.
% Each scene of the two datasets contains physical bottlenecks which are not always leading to congestion situations in pedestrian movements.
% For this reason, the data sets were extended with ground truth data that has both spatial and temporal properties. For this spatio-temporal problem, a new metric is presented which makes the current method and future methods comparable.
%\texttodo{Furthermore, the existing AGORASET and Jülich data sets were updated with Ground Truth data for recognition purposes. A metric is presented for the evaluation of the data.}

% \texttodo{In the work of Solmaz et al.\cite{Solmaz_2012}, four additional crowd scene properties (blocking, lane, ring/arch, fountainhead) are detected in addition to bottleneck situations using optical flow. The method performs well for these properties, but has problems with the superposition of motion patterns.}
In the work of Solmaz et al. \cite{Solmaz_2012}, four additional crowd scene properties (blocking, lane, ring/arch, fountainhead) are detected in addition to bottleneck situations using optical flow. The method performs well for these properties, but has problems with the superposition of motion patterns.

In this work we propose a novel video-based bottleneck detector based on the evaluation of characteristic  stowage patterns in crowd-movements by segmenting the crowd flow.
The idea of this detector is that physical bottlenecks are related to bottlenecks in the contours of the crowd flow segments.
To segment the crowd flow contour we apply long-term analysis based on the Lagrangian framework proposed in \cite{Kuhn_2012} and use the Finite Time Lyapunov Exponents (FTLE) field to extract motion boundaries.
High ridges in the FTLE field indicate Lagrangian features that are assumed to be located at motion boundaries.
In addition we propose a long-term temporal low-pass filtered FTLE to suppress unsteady local features in the Lagrangian field that are caused by heterogeneous motion of the people in the crowd and lead to erroneous crowd flow contours. 

The bottleneck location is defined by the center of a point pair, which is found by geometrical and temporal consistency constrains applied to bottleneck candidates. 
Bottleneck candidates are defined by defects on the contour, i.e. points on the contour with a maximum distance to the contours convex hull. 
To evaluate the performance of the proposed system we manually annotated a selected set of bottleneck sequences from the synthetic AGORASET and the J\"ulich dataset.
\section{Lagrangian Measures for Bottleneck Detection}
\label{sec:Lagrangian_Measures_for_Bottleneck_Detection}
%Lagrangian methods have its origin in the unsteady flow visualization and analysis where it has been proven to be a powerful tool for analyzing computational fluid dynamics for instance to design fluid-dynamic systems.
The origin of Lagrangian methods lies in the visualization and analysis of unsteady flows and has been proven to be a powerful tool for analyzing computational fluid dynamics for instance to design fluid-dynamic systems.
These methods are used to describe non-linear dynamic systems that are represented by a series of time-dependent vector fields. 
The pioneering work by Ali and Shah \cite{Ali_2007} first showed that the Lagrangian methodology can be useful for video-based crowd segmentation. 
Inspired by this work Kuhn \etal proposed in \cite{Kuhn_2012} a compact and applicable framework that implements Lagrangian concepts for video analytics.
%for video analytics by utilizing the Lagrangian concepts and translate them to the video domain.
%
At its core, this framework is based on characterizing motion as a sequence of optical flow fields $\vv(\xx,t)$ to assemble a time-dependent vector field that encodes the dynamics of the video sequence in space-time of a temporal range $[t_0, t_0 + \tau ]$. 
In this work we will follow this framework where the analysis of the optical flow fields is based on so called path lines \cite{Kuhn_2012}.
Path lines can be interpreted as traces of massless particles advected in the flow fields.
Their computation, i.e. advection, is based on the computation of the flow map $\phi^{\tau}_{t_0}(\xx) = \phi(\xx, t_0, \tau)$ which is a core aspect of Lagrangian methods.
The flow map defines the mapping of all massless particles at time $t_0$ seeded at the position $\xx_0$ to their corresponding positions after an integration time $\tau$ :

\begin{equation}
    \phi^{\tau}_{t_0} : D \rightarrow D : \phi^{\tau}_{t_0}(\xx_0) = \xx(t:t_0,\xx_0),
\end{equation}
$t_0$ is the so called frame of reference denoting the basis of the projection of the path line properties.
The flow map $\phi^{\tau}_{t_0}$ is constructed by integrating path lines in
the optical flow fields over $\tau$ time steps, i.e. propagating the massless particles at position $\xx_t$ and time $t$ based on the flow vector $\vv(\xx_t, t)$. 
Since the optical flow fields are discrete in space and time, trilinear interpolation is applied and the particle position is updated by:
\begin{equation}
  \xx_{t+1} = \xx_t + \tilde{\vv}(\xx_t,t),
\end{equation}
where $\tilde{\vv}$ denotes the interpolated motion vectors.
It is assumed that these path lines characterize the overall dynamics, i.e. motions,  and can provide quantitative information about the observed objects in the scene. 
Instead of considering individual trajectories only, this information can be compactly represented within so called Lagrangian fields. 
Examples of Lagrangian fields that have been applied for video analysis are the arc length field \cite{Kuhn_2012} for segmentation or the direction field for violence detection \cite{Senst2017} or action recognition \cite{Acar2012}.
%Lagrangian fields that have been applied for video analysis are the arc length field \cite{Kuhn_2012} for segmentation or the direction field for violence detection \cite{Senst2017} or action recognition \cite{Acar2012}.

One specifically popular type of Lagrangian fields are Finite-Time Lyapunov Exponents (FTLE) which quantify the amount of separation between neighboring path lines.
With respect to features in non-linear dynamic systems high ridges in the FTLE scalar field are assumed to be in close relationship with Lagrangian Coherent Structures \cite{HALLER2011} (regions of maximum change over time) that can serve as basic features to capture and quantify advanced motion patterns \cite{Haller2015}. 
In the video domain FTLE fields have been successfully used in crowd segmentation \cite{Moore2011, Soori2008, Ali_2007}, motion anomaly detection \cite{Wu2010} and person behaviour analysis \cite{senst_2012}.
High ridges are assumed to be in close relationship with motion boundaries of physical objects with respect to small and large entities.
For the task of crowd segmentation these ridges have shown to be salient and stable features.
%
%On those grounds we want utilize FTLE field to extract the distinct shape of crowd moving through a bottleneck.
On those grounds we want utilize FTLE fields to extract the distinct shape of crowds moving through bottlenecks.
The FTLE is derived from the spatial gradients of the flow map.
With
\begin{equation}
	\nabla \phi^{\tau}_{t_0}(\xx) = \frac{\partial \phi^{\tau}_{t_0}(\xx)}{\partial \xx}
\label{eq:flow_map_gradient}
\end{equation}
being the spatial gradients of the flow map and
\begin{equation}
	\mu_i = ln \sqrt{\lambda_i (\nabla^T \nabla)},
\label{eq:flow_map_gradient_eigendvalue}
\end{equation}
the FTLE value for integration time $\tau$ is obtained as 
\begin{equation}
 \text{FTLE}^{\tau}(\xx,t_0) = \frac{1}{\tau} \text{max} \{\mu_1, \mu_2 \}.
\end{equation}
$\nabla^T$ is the transposed of $\nabla$ and $\lambda_i(\nabla^T \nabla)$ denotes the $i$-th eigenvalue of the symmetric matrix $\nabla^T \nabla$.
In general the FTLE can be computed in forward and backward direction resulting in the description of FTLE+ and FTLE-.
The forward FTLE field describes regions of repelling LCS,
while features in the backward FTLE describe attracting LCS structures over the considered time scope.
Only the intersections of FTLE+ and FTLE- ridge structures can segment regions of coherent movement and group invariant moving areas within the motion field. 
%

%\input{section2_old}
% method
\section{Video-based Bottleneck Detection}
% 
%Still \cite{Still_2000} describe that at bottlenecks the pedestrian flow has a characteristic pattern. 
% 
With the proposed method we want to localize the physical bottlenecks by analyzing person flow patterns around narrow places. 
%EB: alternativ: narrow places 
%
This localization will be based on the segmentation of the crowd flow.
We have observed that physical bottlenecks result in bottlenecks in the contour of the crowd flow segments.
We assume that the shape of crowd flow segments can be estimated by extracting high ridges in FTLE fields and propose a long-term analysis of the scene since the movements in that area can be very small.  
So called defects in the contour of the crowd flow shape allow us to detect salient points that restrict the bottleneck. 

Our approach will be composed of three major parts: the long-term temporal filtered FTLE fields, the crowd flow contour segmentation and crowd flow contour analysis.

\begin{figure}
\centering
\scriptsize
\begin{tabular}{c}
  \vspace{0.15cm}
  \includegraphics[width=3.7cm,height=2.4cm]{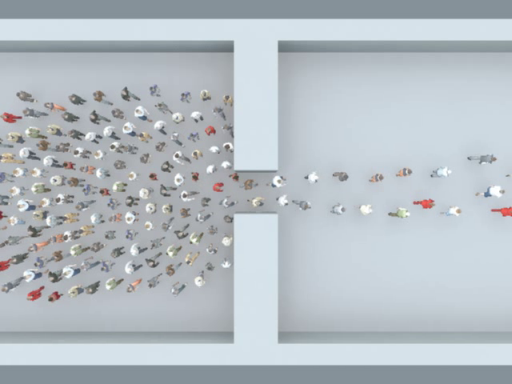} 
  \includegraphics[width=3.7cm,height=2.4cm]{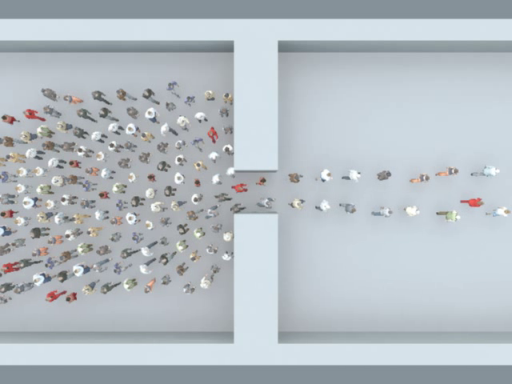}
  \\
  a) AGORASET sequence scene04\_x1\_view1 \\ % AGORASET
  \vspace{0.15cm}
  \includegraphics[width=3.7cm,height=2.4cm]{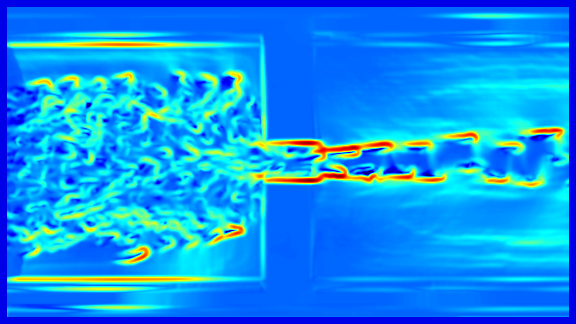} 
  \includegraphics[width=3.7cm,height=2.4cm]{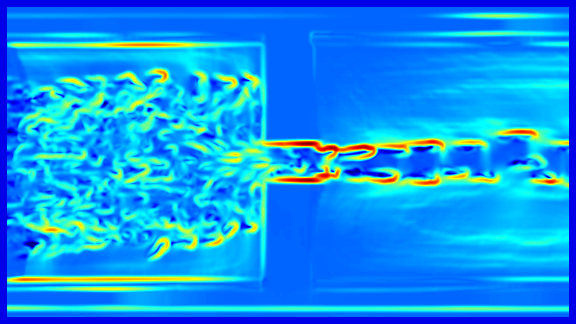} \\
  b) FTLE- \\
  \includegraphics[width=3.7cm,height=2.4cm]{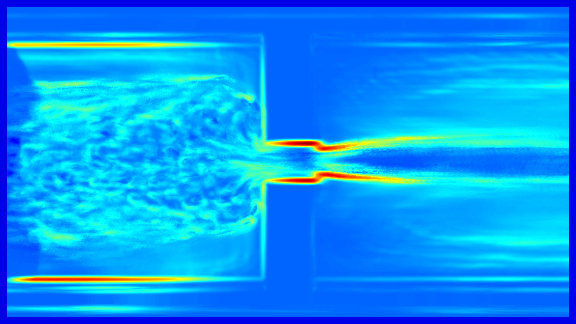} 
  \includegraphics[width=3.7cm,height=2.4cm]{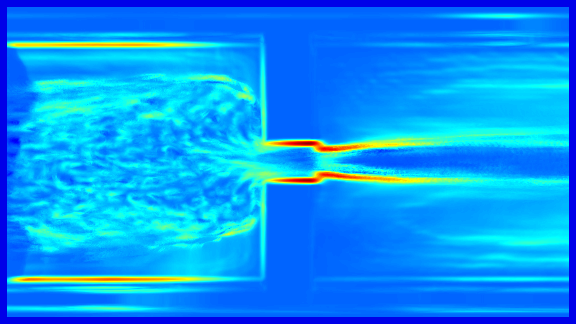} 
  \\
  c) $\tFTLE$- \\
  
\end{tabular}
\caption{Comparison of the FTLE fields estimated in backward direction for $\tau=15$ containing unsteady and long-term temporal filtered FTLE ($\tFTLE$-). In contrast to the $\tFTLE$- the FTLE- contains more locally unsteady structures.}
\label{fig:Long_Term_Filtered_FTLE}
\end{figure}
\textbf{\textit{Long-Term Temporal Filtered FTLE}}: 
For the calculation of the crowd flow shape, a long-term analysis of the scene is of  major interest, since the movements in near the area of the bottleneck can be very small.  
Figure \ref{fig:Long_Term_Filtered_FTLE} (top) shows the FTLE+ field of a rather short integration time $\tau=15$.  
% EB: FTLE- oder FTLE+ in Figure 1 b) ?
%
The ridges related to the crowd margin are relative weak.
To assess the crowd margin while the walking speed of the people is low a large integration interval $\tau > 100$ has to be used, which requires a high computational effort. %EB: very weg?
In addition, due to the heterogeneous movement of the people in the crowd each walking person causes a ridge which becomes stronger for large values of $\tau$. 
%EB: in addition kann eigentlich weg
%
However in contrast to ridges caused by a physical bottleneck and crowd margins, these ridge structures are not consistent for the frame of reference at different times.
%EB: are not consistent when the frame of reference is selected for different times ? kommt darauf an wie es gemeint ist.
%However kann auch weg

To cope with the requirement of this long-term surveillance we propose to skip frames, which simply allows to increase the walking speed of the pedestrian.
% EB: was für surveillance? monitoring?
%we adjust the perceived walking speed by skipping frames...
%
To remove local adverse structures caused by individuals and enhance the global ridge structure of the crowd we propose the long-term temporal filtered FTLE.
% EB: called $\tFTLE$ ?
%
%This allows to use relative small integration intervals and reduces the computational effort while remaining the global separation lines.
This allows to use relative small integration intervals and reduces the computational effort while maintaining the global separation lines.

The long-term temporal filtered FTLE can be estimated as follows. 
%
%At first we subsample the given image sequence $I_t$ with the factor $\Delta t$, where $I_t \in \{I_0, I_{\Delta t} \ldots ,I_{n \cdot \Delta t}\}$ and compute the optical flow fields in forward and backward direction:
At first we subsample the given image sequence $I_t$ with the factor $\Delta t$, where $I_t \in \{I_0, I_{\Delta t} \ldots ,I_{n \cdot \Delta t}\}$ with $n \in \mathbb{N}$ and compute the optical flow fields in forward and backward direction:
\begin{eqnarray}
\vv^+_t \in \{\vv_0(I_0, I_{\Delta t}), \ldots ,\vv_{n \cdot \Delta t} (I_{n \cdot \Delta t}, I_{(n+1) \cdot \Delta t})\} \nonumber \\
\vv^-_t \in \{\vv_0(I_{\Delta t},I_0), \ldots ,\vv_{n \cdot \Delta t} (I_{(n+1) \cdot \Delta t},I_{n \cdot \Delta t})\}. 
\end{eqnarray}
In a next step we compute the $\text{FTLE}_+^{\tau}(\xx,t_0)$ from the optical flow fields $\{\vv^+_{t_0}, \ldots, \vv^+_{t_0 + \tau \cdot \Delta t}\}$ and $\text{FTLE}_-^{\tau}(\xx,t_0)$ respective from $\{\vv^-_{t_0}, \ldots, \vv^-_{t_0 - \tau \cdot \Delta t}\}$ for the reference frame at $t_0 \in \{0,\ldots,n\cdot \Delta t \}$.
Please note that the real integration time is $\tau \cdot \Delta t$. 

For a given set of $\tau_s$ consecutive FTLE fields we apply the temporal low-pass filter using median: %mean or median:
\begin{equation}
%  \underset{mean}{\overline{\text{FTLE}}}^{\tau}(\xx, t_0) &=& \frac{1}{\tau_s} \overset{\tau_s-1}{\underset{n=0}{\sum}} \text{FTLE}^{\tau}(\xx, t_0 + \Delta t \cdot n) \nonumber \\
\overline{\text{FTLE}}^{\tau}(\xx, t_0) = \underset{n \in [0, \tau_s-1]}{\text{median}} \text{FTLE}^{\tau}(\xx, t_0-n\cdot \Delta t). %&=&\text{FTLE}^{\tau}(\xx)_{\frac{\tau_s+1}{2}}.
\end{equation}
% EB: wo ist t_0 rechts?
%
Figure~\ref{fig:Long_Term_Filtered_FTLE} gives an example of the temporal filtered FTLE. 
%EB: Fig 1 c) ?
%
It can be shown that in contrast to the FTLE the $\overline{\text{FTLE}}$ fields are less affected by unsteady, temporal local, ridge structures caused by individual motions and contain features that are steady on a global temporal scale.
% EB: wirklich global oder "more global"? geht doch nur für max \tau und nicht über unendlich lange sequenzen
%

\begin{figure}
\centering
\scriptsize
\begin{tabular}{cc}
  \includegraphics[width=3.75cm,height=2.4cm]{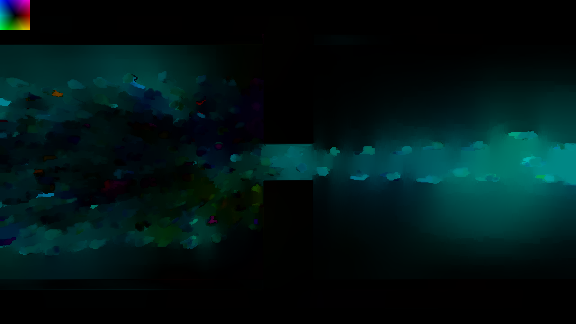} &
  \includegraphics[width=3.75cm,height=2.4cm]{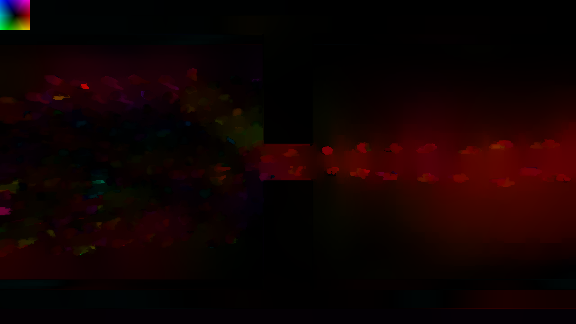} \\
  a) optical flow backward & 
  b) optical flow forward \\
  \includegraphics[width=3.75cm,height=2.4cm]{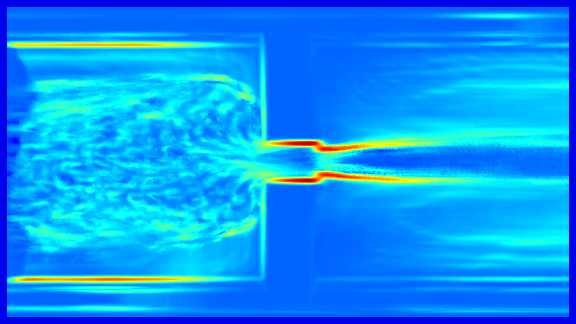} &
  \includegraphics[width=3.75cm,height=2.4cm]{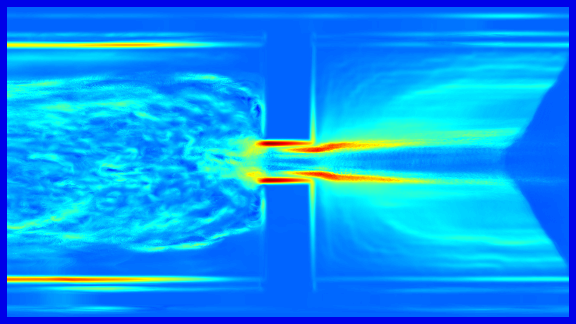} \\
  c) $\tFTLE$- & 
  d) $\tFTLE$+ \\
    \includegraphics[width=3.75cm,height=2.4cm]{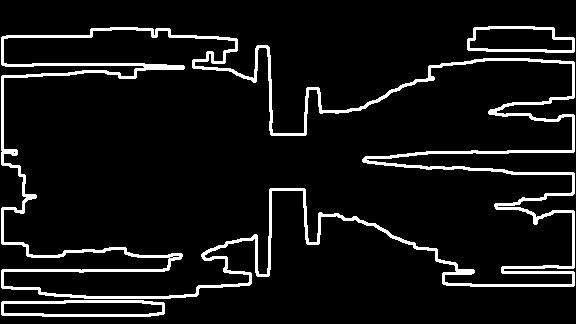} &
  \includegraphics[width=3.75cm,height=2.4cm]{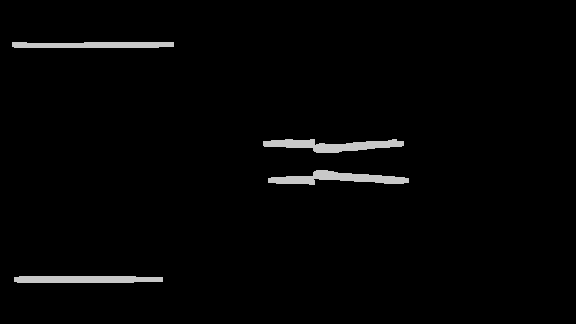} \\
  e) crowd flow contour & 
  f) validation map \\
\end{tabular}
\caption{Illustration of crowd flow contour segmentation. Long-term filtered FTLE fields (c,d) are computed based on RLOF \cite{Senst2016} optical flow (a,b) and lead to the basis of segmentation map (e) and the validation map (f). In (e) the combination of forward and backward ridges in the $\tFTLE$ fields leads to the crowd flow segment. Only the dominant ridges with high values that occurs in both $\tFTLE$ fields are in (f).}
\label{fig:Countour_Segmentation}
\end{figure}
\textbf{\textit{Crowd Flow Contour Segmentation}}: 
Figure \ref{fig:Countour_Segmentation} shows exemplary the extraction of the salient motion boundary contour caused by the crowd flow.
We extract ridges with high and low FTLE values.
The low ridge contour will be the basis to generate possible bottleneck candidates while the high ridge contour will be used to evaluate the bottleneck candidates.
Both are computed by the binarization of the temporal filtered FTLE fields based on the two thresholds $\sigma_{low}$ and $\sigma_{high}$ :
\begin{equation}
  M^{(+-)}_{(low/high)} = \left\{ \xx \mid \tFTLE^{\tau}_{(+-)}(\xx) > \sigma_{(low/high)} \right\}.
\end{equation}
% solltie hier nicht \tFTLE^{\tau}_{(+/-)} statt \tFTLE^{\tau}_{(+-)} stehen?

%
%After dilation all four binary maps via morphological operation to close gaps in the contour a segmentation map $M_{seg}(\xx, t_0) = M^-_{low}(\xx, t_0) \lor M^+_{low}(\xx, t_0)$ is computed by combining the forward and backward low ridge maps.
After dilating all four binary maps to close gaps in the contours a segmentation map $M_{seg}(\xx, t_0) = M^-_{low}(\xx, t_0) \lor M^+_{low}(\xx, t_0)$ is computed by combining the forward and backward low ridge maps.
The segmentation map contains the contour of the crowd flow and can be prone to oversegmentation and artifacts.
A second validation map $M_{val}(\xx, t_0) = M^-_{high}(\xx, t_0) \land M^+_{high}(\xx, t_0)$ will be computed by the overlap of the forward and backward high ridge maps that contains the most stable ridges of the Lagrangian fields. 
%
%These ridges of this map are in relationship to the barriers of the physical bottleneck.
The ridges of this map relate to the barriers of the physical bottleneck.
%
% Unless this map can not contain the complete crowd contour it contains ridges, which are at the bottleneck location with a high probability.
Unless this map can not contain the complete crowd contour it contains ridges that are at the bottleneck location with a high probability.
% EB: dieser Satz ergibt für mich leider keinen Sinn (wurden hier mehrere Sätze unglücklich zusammengesteckt?)
% 

\begin{figure}
\centering
\scriptsize
\begin{tabular}{cc}
  \includegraphics[width=3.75cm,height=2.4cm]{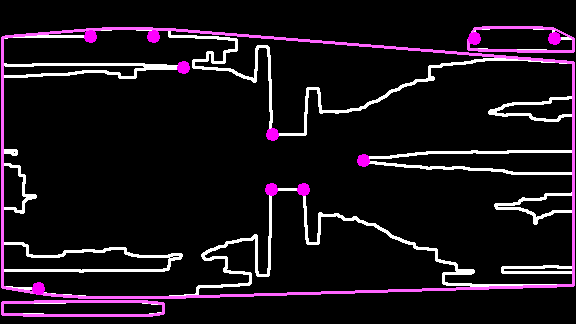} &
  \includegraphics[width=3.75cm,height=2.4cm]{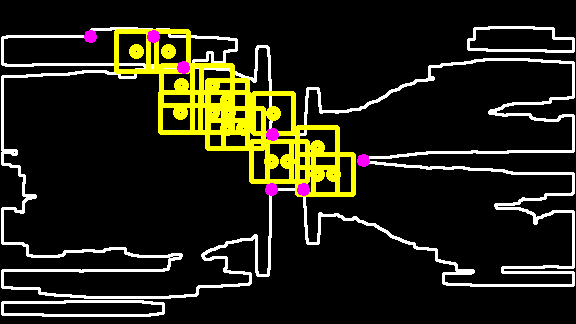} \\
  a) defects and convex hull &% crowd flow contour & 
  b) bottlenecks candidates \\
  \includegraphics[width=3.75cm,height=2.4cm]{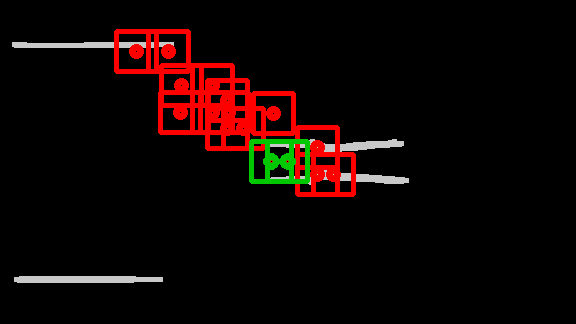} &
  \includegraphics[width=3.75cm,height=2.4cm]{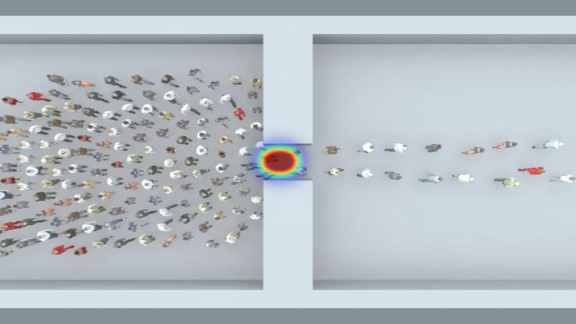} \\
  c) validated bottlenecks & 
  d) density visualisation \\
\end{tabular}
\caption{Illustration of the crowd flow contour analysis for bottleneck detection. Bottleneck candidates are selected by comparing the contour with convex hull (a). Selected candidates (b) are filtered by geometrical filter and the validation map (c). The visual result can be seen in (d).}
\label{fig:Countour_Analysis}
\end{figure}
\textbf{\textit{Crowd Flow Contour Analysis}}: 
An example of the crowd flow contour is shown in Figure~\ref{fig:Countour_Analysis}(a).
The bottleneck candidates are a set of point tuples $\grave{C}_{t_0} = \{(\grave{\xx}_0, \grave{\xx}_1)^0_{t_0}, \ldots, (\grave{\xx}_0, \grave{\xx}_1)_{t_0}^{m} \}$ that are located at indentations of the contour (purple dots). 
The candidates can be located by computing so called defects.
Defects are points computed by evaluating the distance between the contour at its convex hull \cite{Sklansky1982}.
Bottleneck candidates are filtered by two geometrical constraints between the point pair $(\grave{\xx}_0, \grave{\xx}_1)_{t_0}^{m}$: 

i) The relation $d_c / l_s$ between euclidean distance $d_c = ||(\grave{\xx}_0, \grave{\xx}_1)||$ and the crowd flow segment contour length $l_s$ has to be below a given threshold $\sigma_s$ (see Figure ~\ref{fig:Countour_Analysis}(b)).
ii) The relation between the distance of the points on the contour and the euclidean distance has to be greater than $ 2 \cdot d_c $ to remove point pairs that are likely to be not on opposite location of the contour.
% EB: da genug platz übrig ist könnte man hier auch eine aufzählung draus machen
%
Finally, the points are projected on the validation map $M_{val}(\xx,t_0)$.
If the point, as shown in Figure ~\ref{fig:Countour_Analysis}(c), contains at least two different ridges within a region of size $\sigma_r \times \sigma_r$, which are not on the same contour, the center point is selected as a bottleneck detection $c^m_{t_0}=(\xx_0, \xx_1)_{t_0}^{m}$. 
% If the point, as shown in Fig.~\ref{fig:Countour_Analysis}(c), within a rectangle region of $\sigma_r \times \sigma_r$ that is located at their center are not on the same contour, the center point is selected as bottleneck detection $c^m_{t_0}=\{\xx_0, \xx_1\}_{t_0}^{m}$. 
%

%The detection of the candidates can be affected by small changes of the ridge segmentation and result into miss detection.
The detection of candidates can be affected by small changes of the ridge segmentation which can result in a missed detection.
To remain consistent over time the detection points $(\xx_0, \xx_1 )_{t_0}$ will be propagated to the next frame. 
% EB: jede dedetktion wird genau einmal propagiert? die propagierten detektionen aber nicht? bitte konkretisieren
% 
%In addition the detection $c^m_{t_0}$ will be finally accepted if it has been detected along $\sigma_{o}\cdot \Delta t$ frames within the same radius $\sigma_r$.
The detection $c^m_{t_0}$ will only be accepted if it has been detected along $\sigma_{o}\cdot \Delta t$ frames within the same radius $\sigma_r$.

%\input{section3_old}
% evaluation
\section{Evaluation}
\begin{figure}
\centering
\scriptsize
\begin{tabular}{c}
  \includegraphics[width=3.7cm,height=2.4cm]{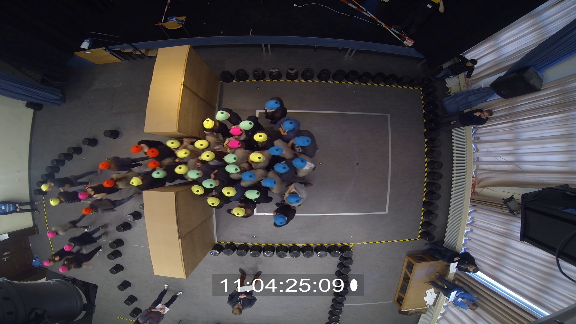} 
  \includegraphics[width=3.7cm,height=2.4cm]{images/dataset/school_GymBay_01_05_170.png} \\
    \includegraphics[width=3.7cm,height=2.4cm]{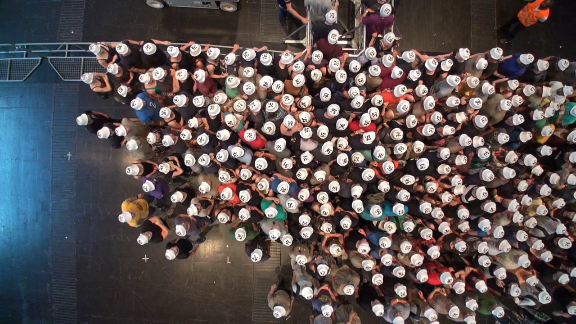} 
  \includegraphics[width=3.7cm,height=2.4cm]{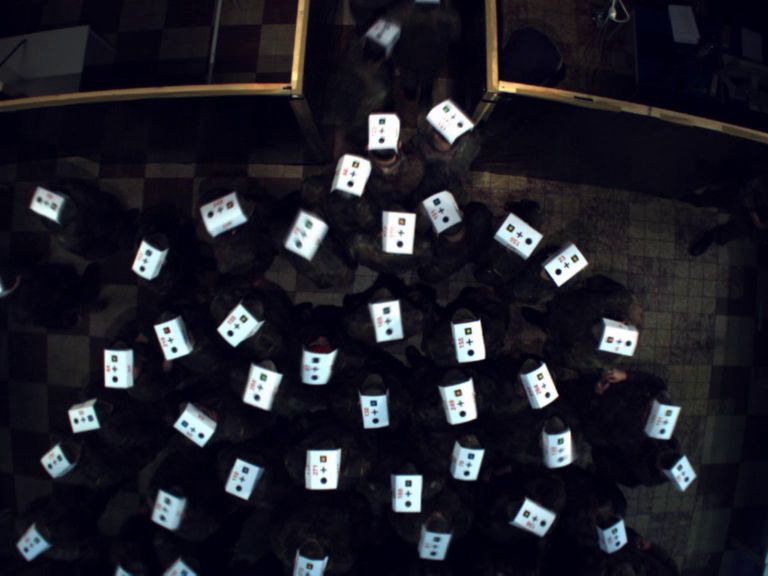} \\
  J\"ulich dataset \\
    \includegraphics[width=3.7cm,height=2.4cm]{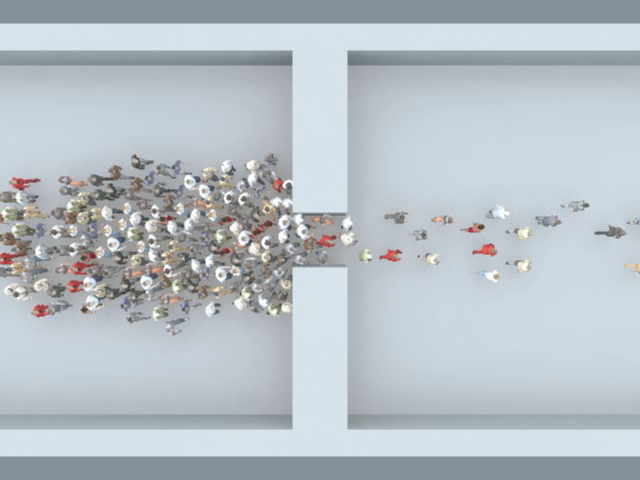} 
  \includegraphics[width=3.7cm,height=2.4cm]{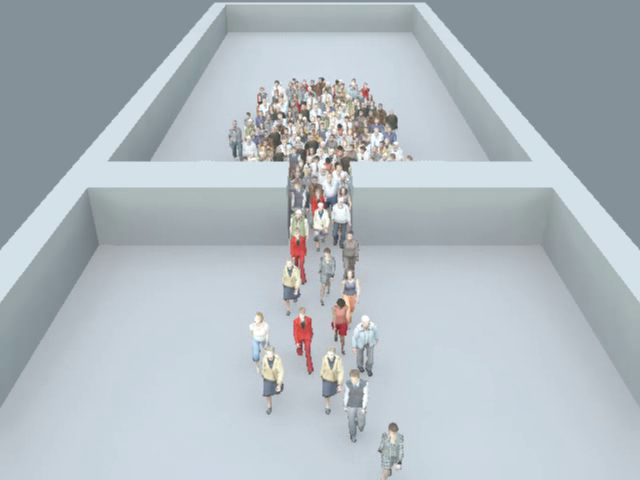} \\
  AGORASET \\
\end{tabular}
\caption{Example of the J\"ulich dataset containing bottlenecks and the synthetic AGORASET.}
%The bottleneck detection method is evaluated by partial sequences of the AGORASET and the pedestrian dynamic archive data set of the Forschungszentrum J\"ulich.}
\label{fig:dataset}
\end{figure}

% \begin{figure}[Ht]%[h]
% \centering
% \begin{subfigure}[hbt]{4.1cm} % 5 
% \begin{center}
% \noindent
%   \includegraphics[width=4.1cm]{images/dataset/school_GymBay_01_05_170.png}
%   \subcaption{\footnotesize{entrance 1}}
%   \label{fig:dataset_entrance}
% \end{center}
% \end{subfigure}
% \hfill
% \begin{subfigure}[hbt]{4.1cm} % 2
% \begin{center}
% \noindent
%   \includegraphics[width=4.1cm]{images/dataset/school_WDG_03b_.png}
%   \subcaption{\footnotesize{
% bottleneck and social groups}}
%   \label{fig:dataset_school_GymBay_01_05}
% \end{center}
% \end{subfigure}
% \hfill
% \begin{subfigure}[hbt]{4.1cm} % 5 
% \begin{center}
% \noindent
%   \includegraphics[width=4.1cm]{images/dataset/scene04_x4_view1_000200.png}
%   \subcaption{\footnotesize{AGORASET scene04\_x4\_view1 }}
%   \label{fig:dataset_view1}
% \end{center}
% \end{subfigure}
% \hfill
% \begin{subfigure}[hbt]{4.1cm} % 2
% \begin{center}
% \noindent
%   \includegraphics[width=4.1cm]{images/dataset/scene04_x4_view2_000200.png}
%   \subcaption{\footnotesize{AGORASET scene04\_x4\_view2}}
%   \label{fig:dataset_ue}
% \end{center}
% \end{subfigure}
% \caption{\texttodo{Dataset caption}}
% \label{fig:dataset}
% \end{figure}
In this section, we assess the performance of the Lagrangian-based bottleneck detection approach.
To evaluate the results we introduce an appropriate metric for this novel problem and provide supplemented ground truth for the existing datasets J\"ulich and AGORASET which are applicable to the new use case.
The ground truth as well as the evaluation script are publicly available for future work\footnote{\scriptsize{https://github.com/simonmaik/bottleneck-detection-benchmark}}.

%The ground truth as well as the evaluation script are publicly available for other work\footnote{\scriptsize{Added upon acceptance of the paper}}.

\begin{figure*}
\centering
\scriptsize
\begin{tabular}{ccc}
    & All sequences & \\
  \includegraphics[width=5.3cm, height=3.9cm]{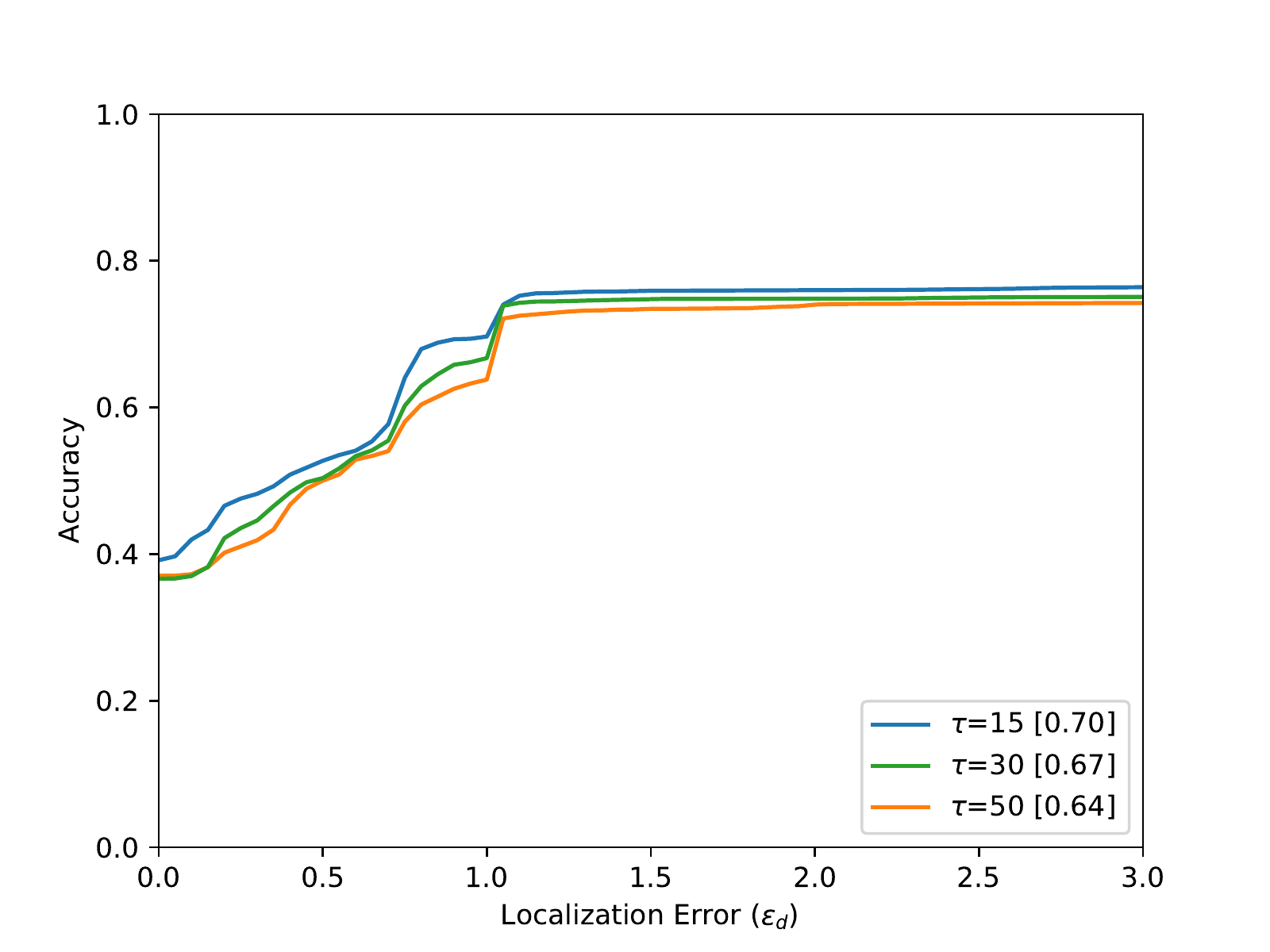} &
  \includegraphics[width=5.3cm, height=3.9cm]{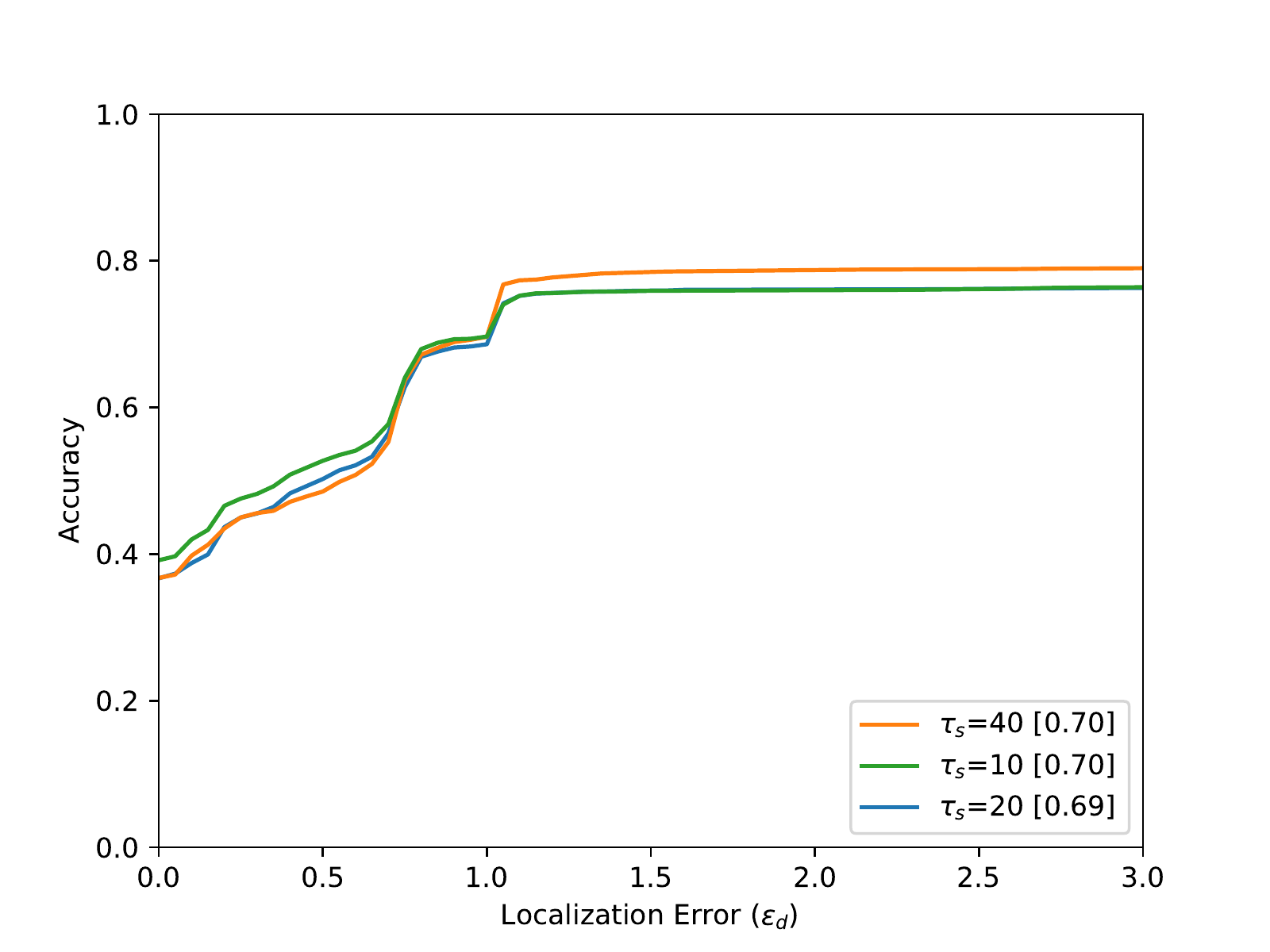} &
  \includegraphics[width=5.3cm, height=3.9cm]{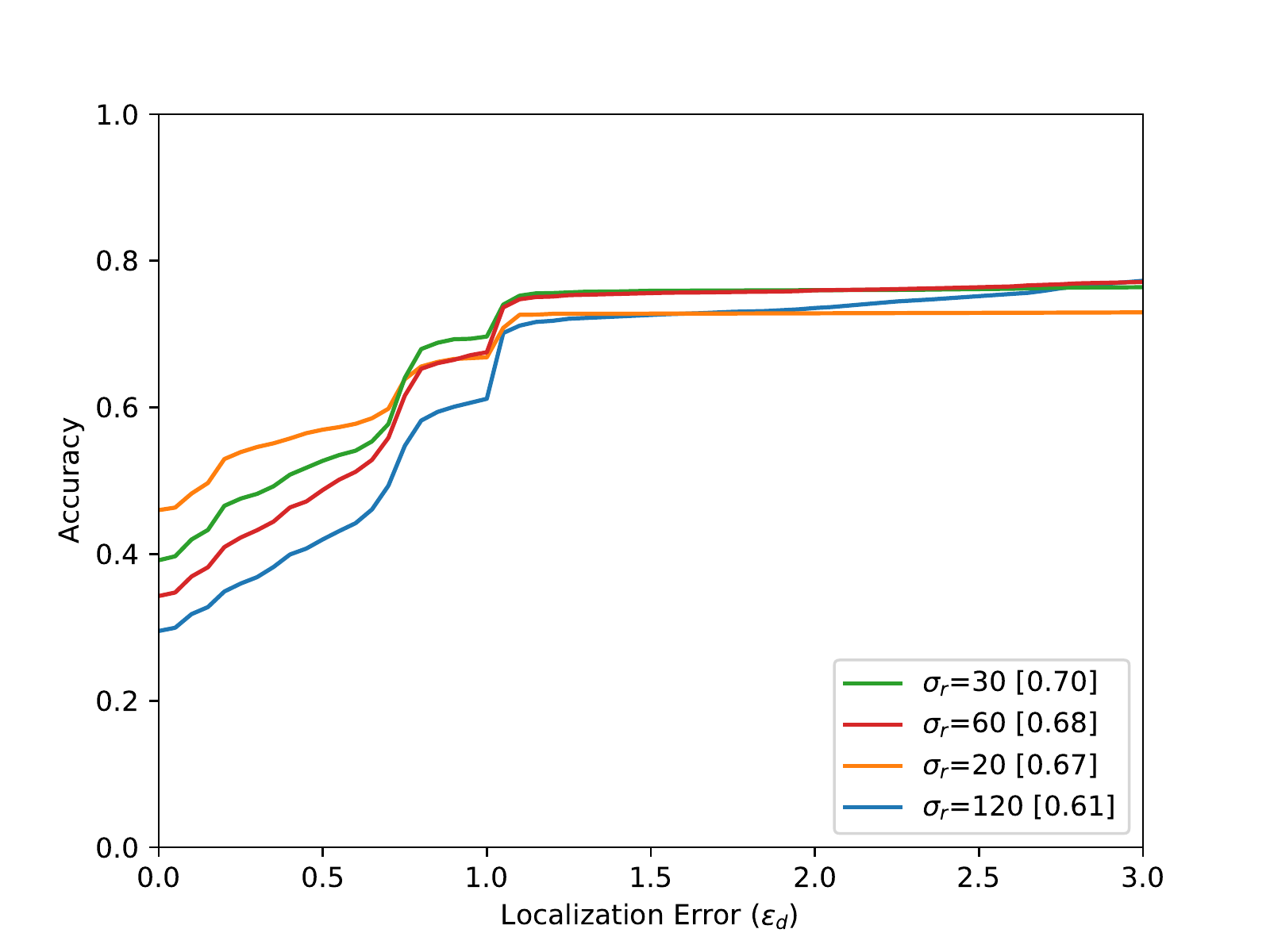} \\
  %a) variation of $\tau$ & b) variation of $\tau_{s}$ & c) variation of $ \sigma_{r} $ 
  \\ 
   & Bottleneck and social groups  & \\
  \includegraphics[width=5.3cm, height=3.9cm]{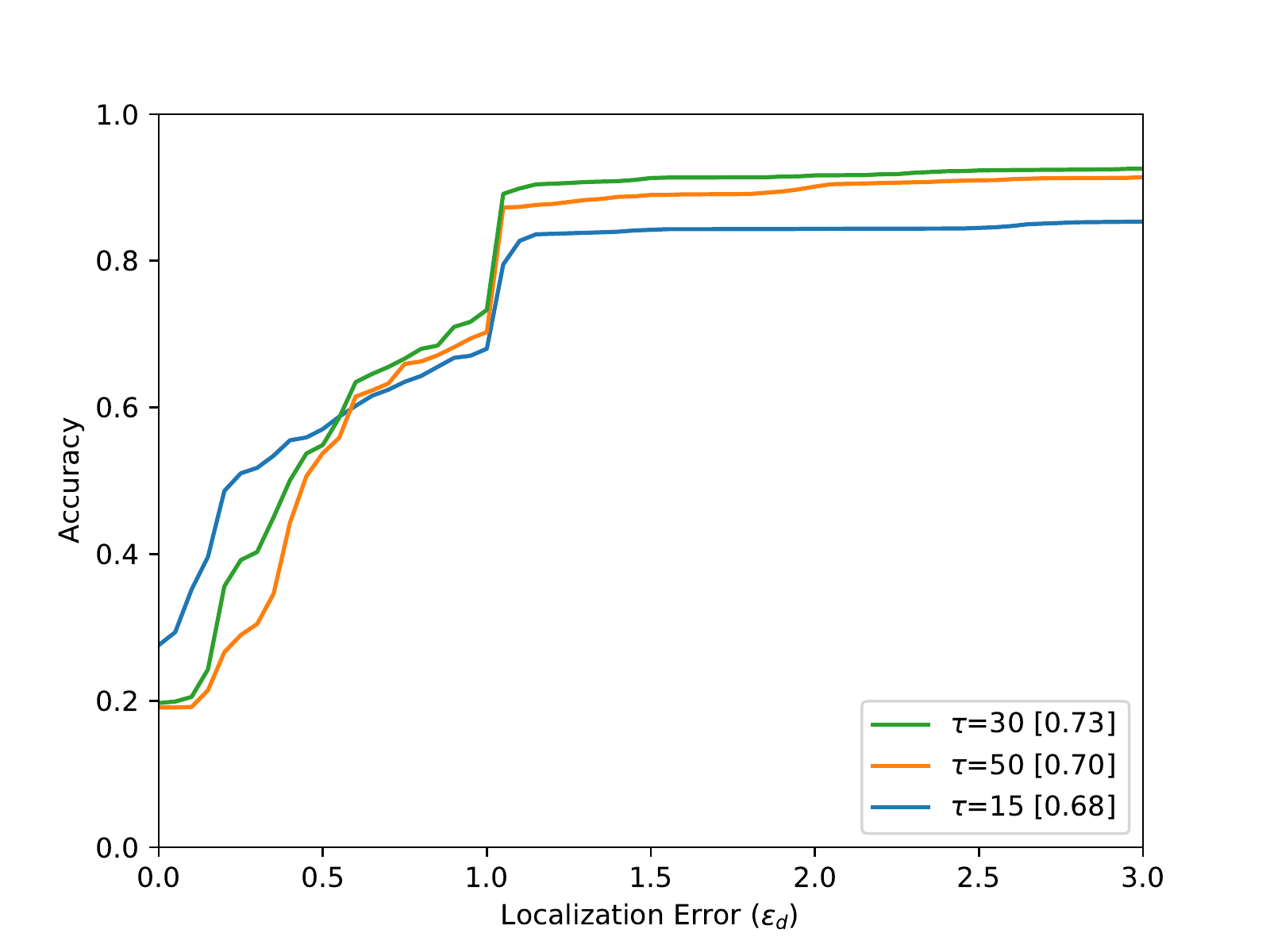} &
  \includegraphics[width=5.3cm, height=3.9cm]{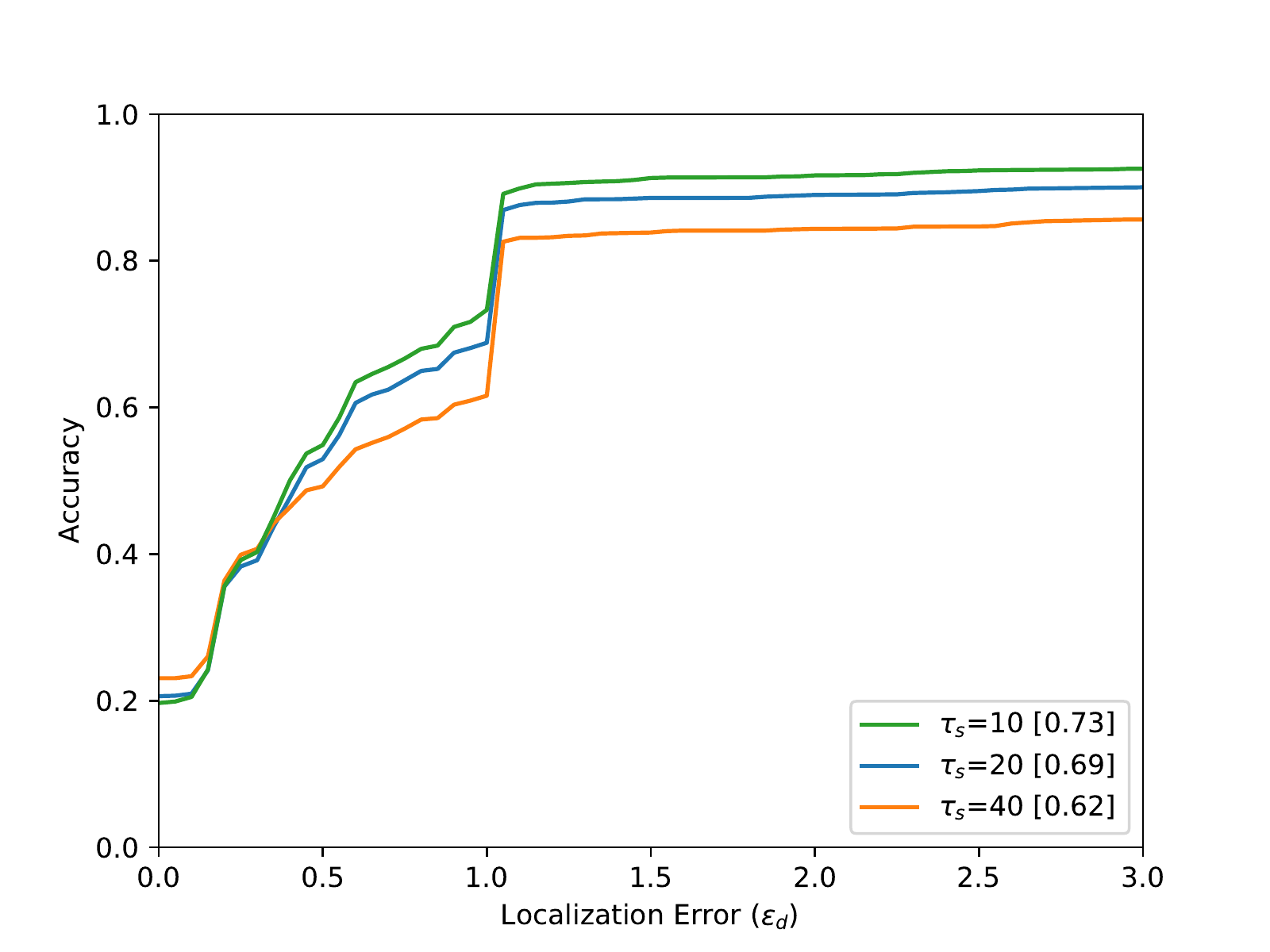} &
  \includegraphics[width=5.3cm, height=3.9cm]{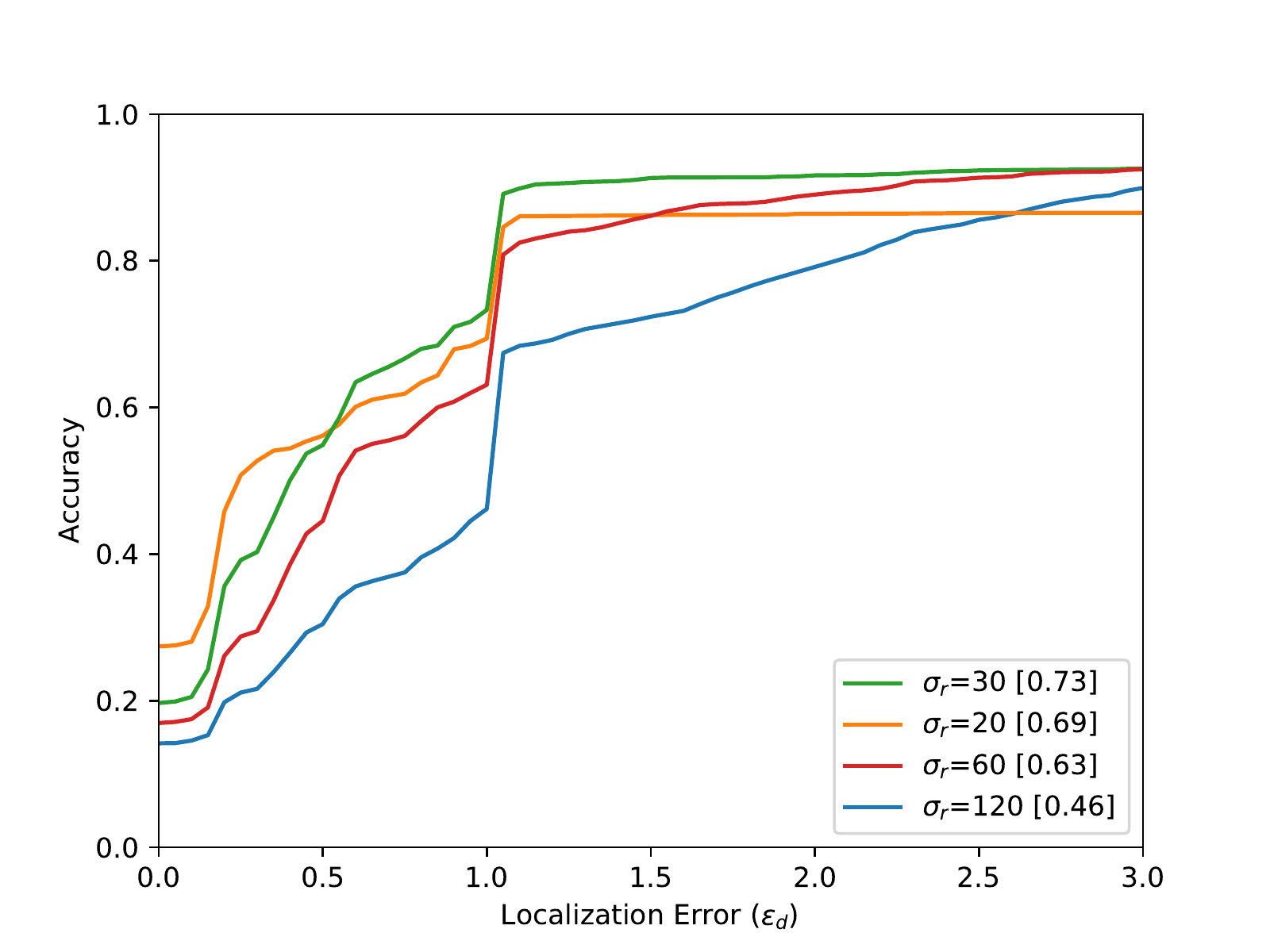} \\
  %d) variation of $\tau$ & e) variation of $\tau_{s}$ & f) variation of $ \sigma_{r} $ 
  \\\\
   & Entrance 1, entry without guiding barriers (semicircle setup) & \\
  \includegraphics[width=5.3cm, height=3.9cm]{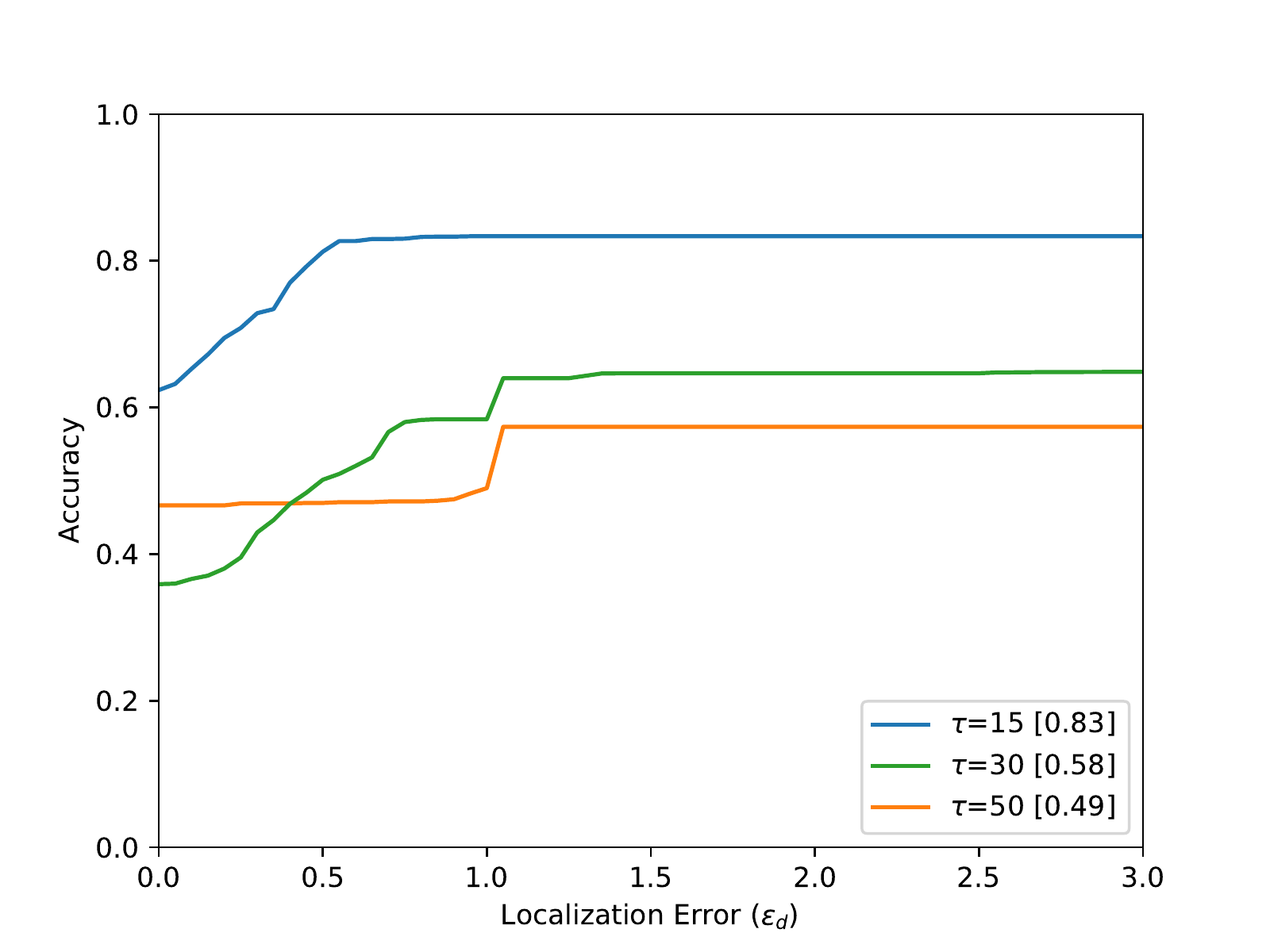} &
  \includegraphics[width=5.3cm, height=3.9cm]{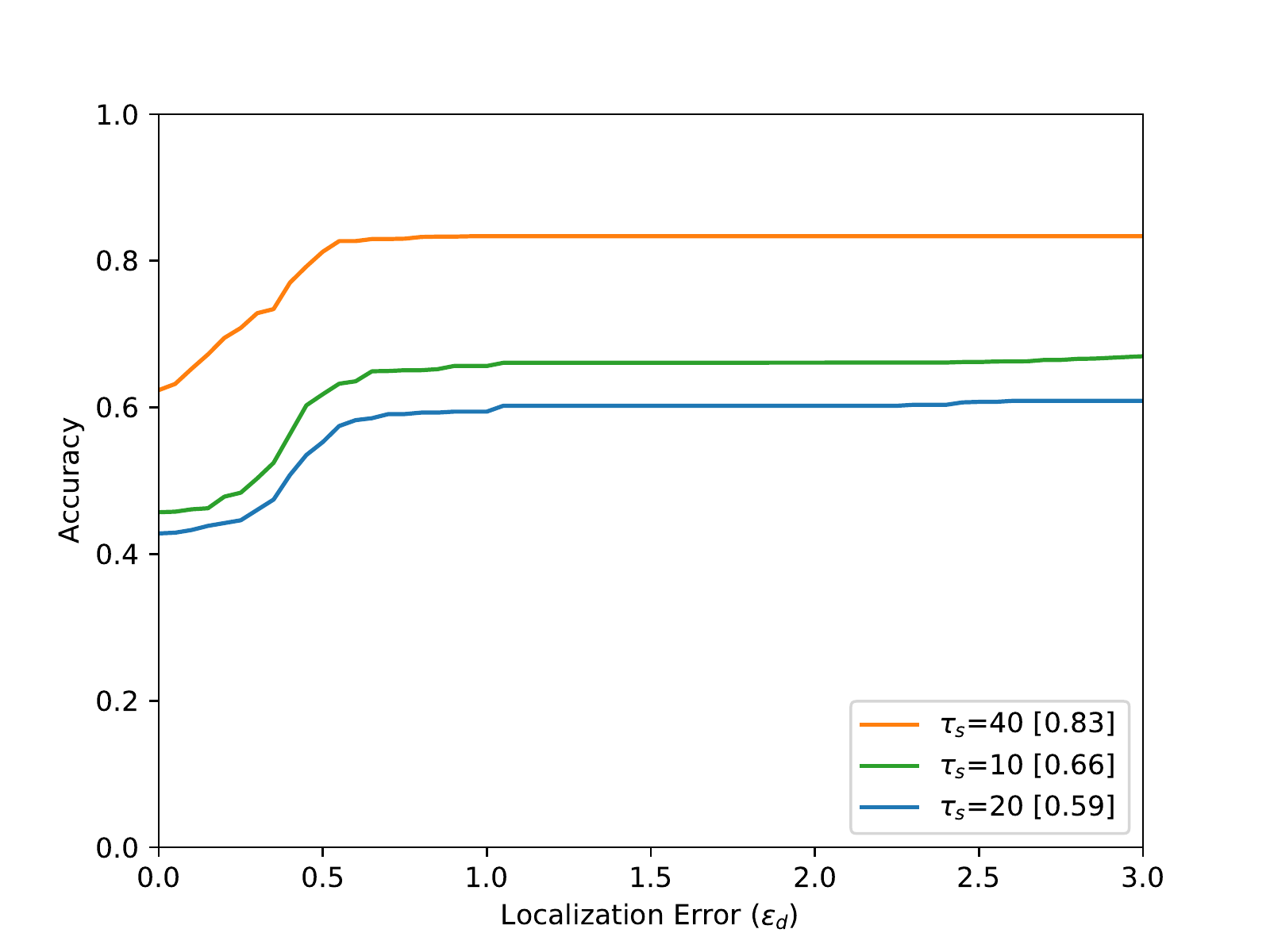} &
  \includegraphics[width=5.3cm, height=3.9cm]{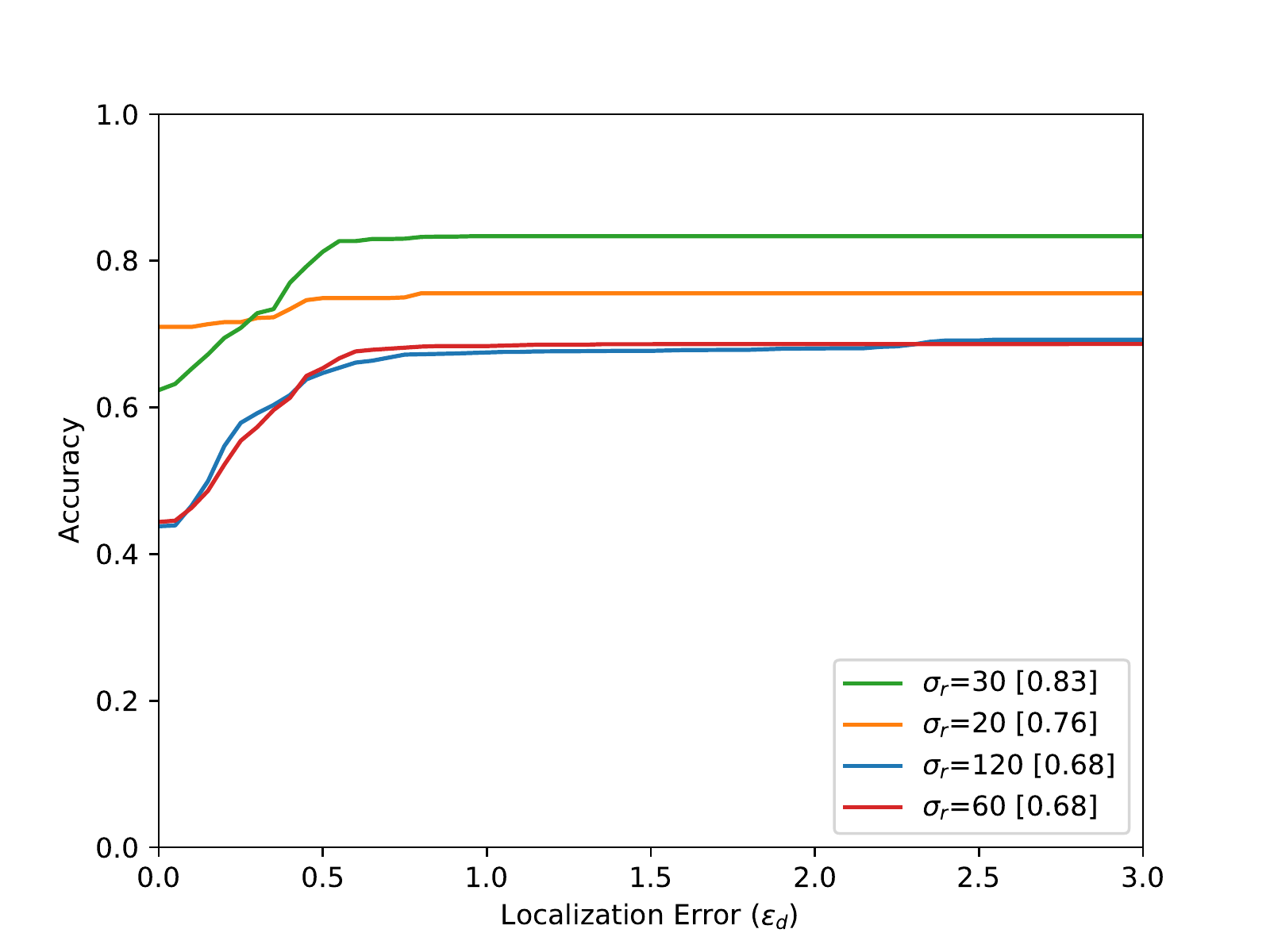} \\
  %g) variation of $\tau$ & h) variation of $\tau_{s}$ & i) variation of $ \sigma_{r} $ 
  \\  \\
     & AGORASET escape sequences & \\
  \includegraphics[width=5.3cm, height=3.9cm]{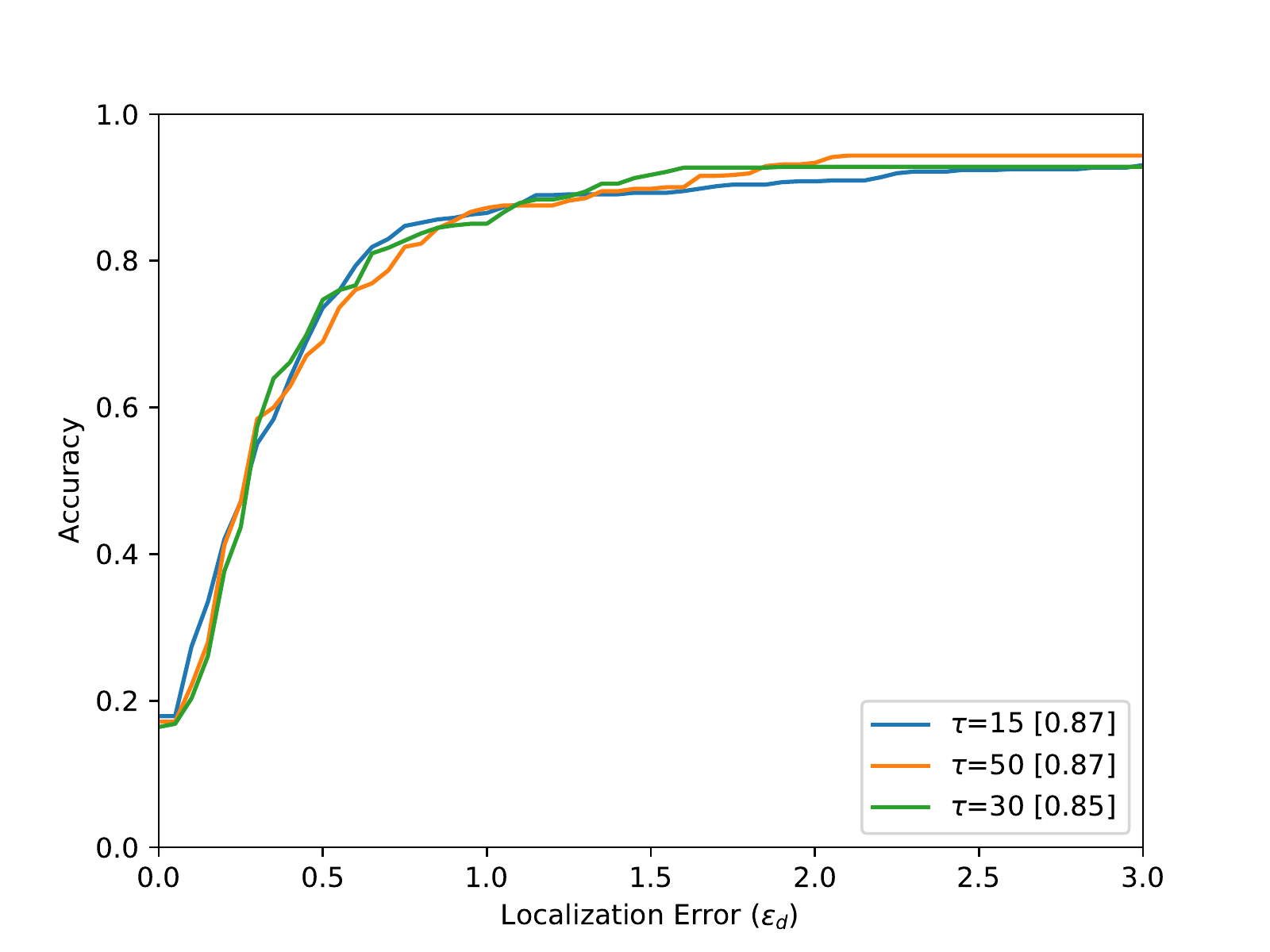} &
  \includegraphics[width=5.3cm, height=3.9cm]{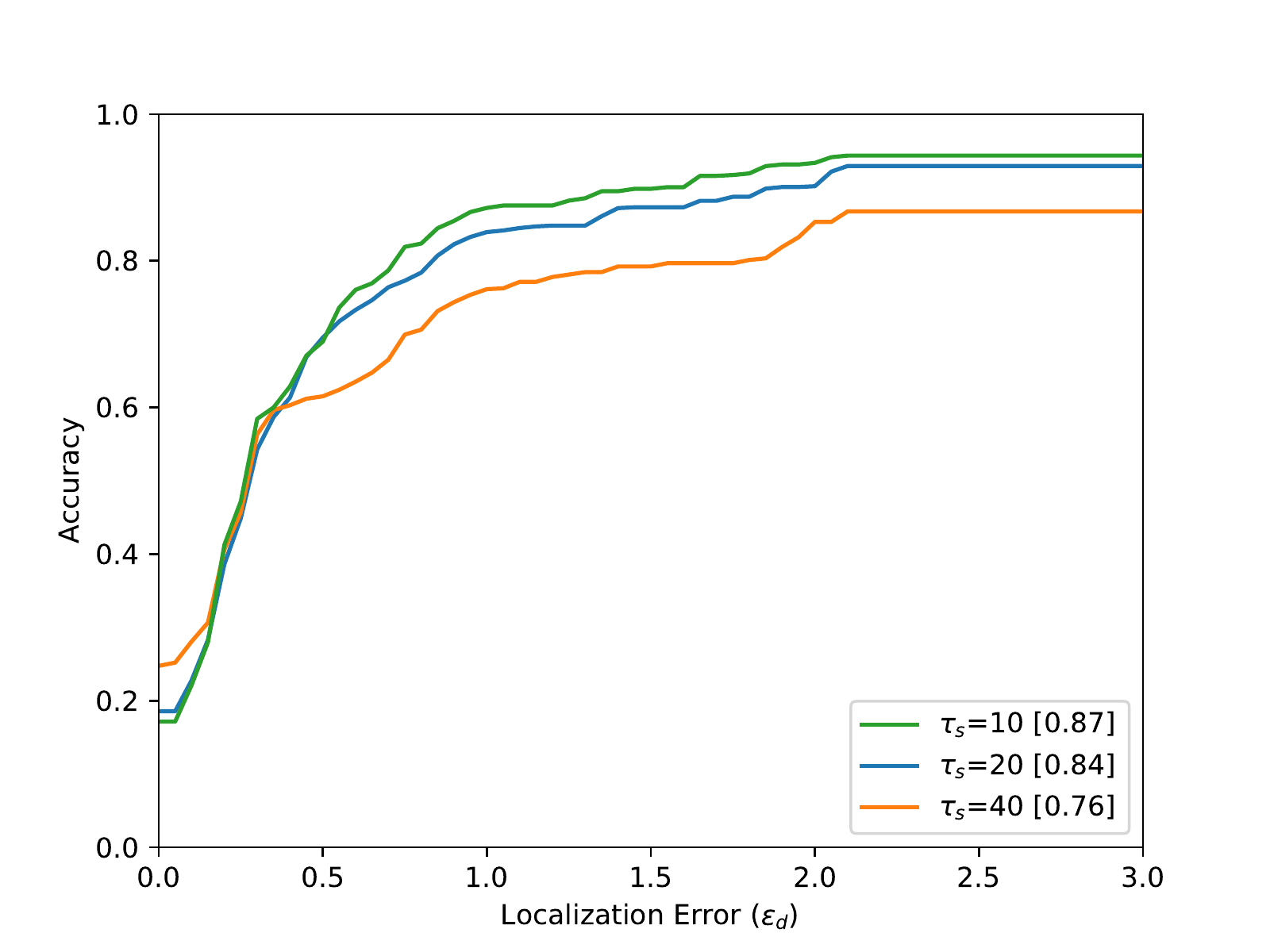} &
  \includegraphics[width=5.3cm, height=3.9cm]{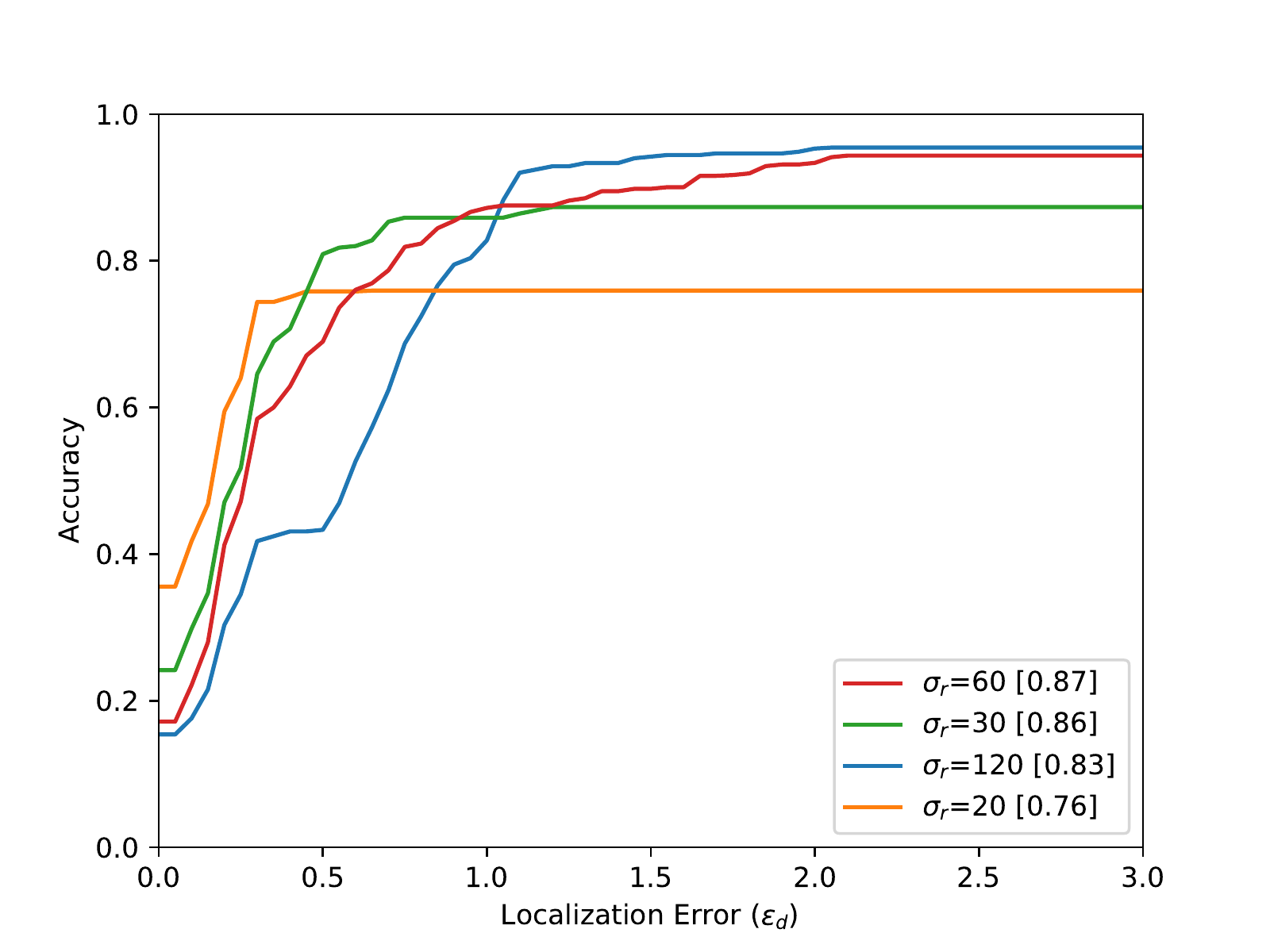} \\
 % j) variation of $\tau$ & k) variation of $\tau_{s}$ & l) variation of $ \sigma_{r} $ 
  \\  
\end{tabular}
\caption{The figure shows the results of the bottleneck detector related to the accuracy value and the localization error. The arrangement from left to right shows the results of the parameter tests: integration time $\tau$ (left), buffer size $\tau_s$ (middle) and radius $\sigma_r$ (right). The plots at the top show the results for all sequences, below are the results for the sequences of the subsets \textit{Bottleneck and social groups} and the single sequence \textit{Entrance 1, entry without guiding barriers (semicircle setup)} . The results for the four escape sequences of the AGORASET dataset are shown at the bottom of this figure. The order within the legends arranges the results after the accuracy at the position $\epsilon_d = 1$. The accuracy value at this position is given in brackets.}
% eignetlich muss man die sortierung auch nicht erklären

\label{fig:results_1}
\end{figure*}

% In this section, we assess the performance of the Lagrangian-based bottleneck detection approach.
%
%To evaluate the results we introduce an appropriate metric for this novel problem and provide ground truth for the pre-existing datasets J\"ulich and AGORASET which are applicable to the new use case so that future research can use our method and evaluation protocol for reference.
%The evaluation will be based on 76 selected sequences from the J\"ulich dataset and four from the AGORASET\footnote{\scriptsize{https://www.sites.univ-rennes2.fr/costel/corpetti/agoraset/Site/Scenes.html}}  \cite{allain_2012} showing crowds passing bottleneck scenarios.
The evaluation will be based on 76 sequences from the J\"ulich dataset and four from the AGORASET\footnote{\scriptsize{https://www.sites.univ-rennes2.fr/costel/corpetti/agoraset/Site/Scenes.html}}  \cite{allain_2012} showing crowds passing bottleneck scenarios.
%
%
%
%
% EB: ohne selected, das klingt sonst so geschummelt
% crowds passing bottleneck scenarios oder lieber irgenwas wie crowds in bottleneck scenarios etc
%
An example of the used sequences can be found in Figure \ref{fig:dataset}. 
%The described method was extensively evaluated for eighty sequences %from the datasets AGORASET and J\"ulich. 
AGORASET is a synthetic rendered dataset. 
It contains different viewing angles as well as a high variation of the peoples density and movement characteristics under constant environmental conditions. 
%
%The four escape sequences%
%\footnote{\scriptsize{https://www.sites.univ-rennes2.fr/costel/corpetti/agoraset/Site/Scenes.html}} 
%are used for our testing.
%
The J\"ulich dataset is a composition of datasets related to \cite{Liao_2014,seyfried_2009,sieben_2017,kruechten_2016} and published 
%\footnote{\scriptsize{http://ped.fz-juelich.de/da/2013entrSemicircle}}%
%\footnote{\scriptsize{http://ped.fz-juelich.de/da/2013entrCorridor}}
%\footnote{\scriptsize{http://ped.fz-juelich.de/da/2009bottleneck}}%
%\footnote{\scriptsize{http://ped.fz-juelich.de/da/2006bottleneck}} 
via the pedestrian dynamics data  archive\footnote{\scriptsize{http://ped.fz-juelich.de/da/}}. 
Different bottleneck sizes were examined as well as different social aspects and their consequences under laboratory conditions.
This has resulted in a broad field of data in which no bottleneck is available for longer periods of time or in which different motion sequences repeatedly occur due to constrictions of varying sizes.
The presented algorithm detects a bottleneck both temporally and spatially.
%
% For temporal evaluation and to measure the accuracy of our detections we annotated ground truth.
%Since there is no ground truth for our consideration, we created one. 
% Den Satz hier am besten weg, die gt kommt später noch. Von Agoraset sagst du aj auch nix zur GT
%
The temporal characteristics of the event were determined under two essential aspects:
i) Pedestrians cross the bottleneck and
ii) the individuals of the crowd try to take the shortest route,  which depends on the density of the crowd's dependent speed.
The last mentioned aspect is based on hypotheses describing a crowd, more details can be found in the work of Hughes \cite{hughes_2003}.
%EB ..based on hypotheses of how to describe a crowd, ... ?
%
%The characterization is necessary, since a narrowing for individual persons in our understanding does not represent a bottleneck.
The characterization is necessary, since a narrowing of individual persons in our understanding does not represent a bottleneck.
% EB: soll gesagt werden das personen enger und langsamer werden müssen? das kommt nicht so richtig raus
%
%For the determination of the spatial ground truth, the central point of the constriction was determined, which was created after the subjective evaluation by scientific staff. 
\begin{figure}
	\centering
	\scriptsize
	\includegraphics[width=0.99\columnwidth]{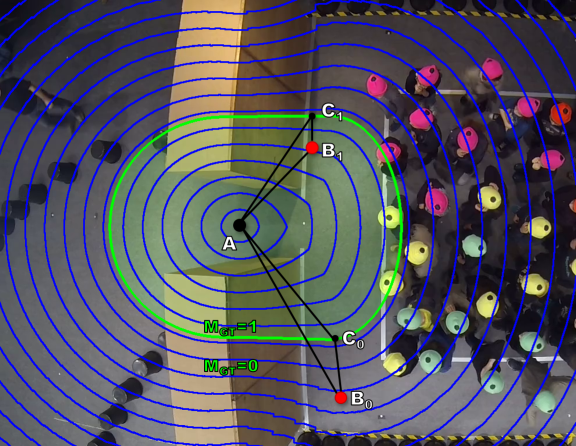}
	\caption{Visualization of the localization error $\epsilon_d$ estimation, the ground truth bottleneck detection ($A$) and the ground truth bottleneck mask $M_{GT}$ (framed by green line).
	$\epsilon_d$ is computed by the relation between the distances between the ground truth detection ($A$), the estimated detection's ($B_0$,$B_1$) and the nearest point to the bottleneck mask ($C_0$,$C_1$).
	Each blue line shows the increasing isometric $\epsilon_d$.}
	\label{fig:description_metric}
\end{figure}
The central point of the constriction was determined and carefully annotated after the subjective evaluation by scientific staff. 
%
%In order to measure accuracy and to take distortion, camera viewing angle and height into account, a binary ground truth map $M_{GT}$ is created around the determined point.
In order to measure the accuracy while taking the distortion, camera angle and scaling/height into account, a binary ground truth map $M_{GT}$ is created around the determined point.
Whereby $M_{GT}$ only exists at the $t$ points in time when a bottleneck exists according to the above time definition.
%
%
%For the calculation of the accuracy it is therefore possible to perform a binary classification:
In our evaluation, the detection of a bottleneck is then treated as a frame-wise binary classification problem with:
True positives $TP$ (mask hit), false positives $FP$ (detection outside mask), true negatives $TN$ (no detection outside mask), false negatives $FN$ (no detection inside mask).
The accuracy is defined as follows:
\begin{equation}
    \text{Accuracy} = \frac{TP + TN}{TP + TN + FP + FN}.
\end{equation}
\begin{table}
    \footnotesize%scriptsize
    \centering
    \begin{tabular}{|c|c|}
    \hline
        {Test set}& {Accuracy} \\
        \hline
      {All sequences }                &      {$0.70$}\\
      {Bottleneck and social groups}  &       {$0.73$}\\
      {Entrance 1             }       &       {$0.83$}\\
      {AGORASET               }       &       {$0.87$} \\
      \hline
    \end{tabular}
    \caption{The Accuracy values for the considered sets shown in Figure \ref{fig:results_1} are listed for localization error values $\epsilon_d = 1$.}
    \label{tab:my_label}
    \vspace{-0.3cm}
\end{table}
In order to evaluate the detection spatially, an additional isometric score $\epsilon_d$, called localization error, is mapped to simulate the dilatation/contraction of $M_{GT}$:
%
% \begin{equation}
%     \epsilon_d = \begin{cases} 
%     \frac{-b}{a+b} + 1\hspace{0.5cm} \text{ if $c^m_{t{_0}}$  }\in M_t\\ 
%     \frac{b}{a}+1 \hspace{0.85cm}\text{ if $c^m_{t{_0}}$  } \notin M_t
%     \end{cases}
% \end{equation}
%
% \begin{equation}
%     \epsilon_d = \begin{cases} 
%     \frac{a}{a+b}\hspace{1.1cm} \text{ if $c^m_{t{_0}}$  }\in M_t\\ 
%     \frac{b}{a}+1 \hspace{0.85cm}\text{ if $c^m_{t{_0}}$  } \notin M_t
%     \end{cases}
% \end{equation}
%
\begin{equation}
    \epsilon_d = \begin{cases}
    \frac{\overline{AB}}{\overline{AB} + \overline{BC}}
    &,if \  M_{GT}(c^m_{t{_0}}) = 1
    \\
    \\
    %\frac{\overline{BC}}{\overline{AC}}+1 &, else%\text{ if } 
    \frac{\overline{BC}}{\overline{AC}}+1 &, else.
    \end{cases}
\end{equation}
% Mit: A GT Punkt; B detektierter Punkt; C n\"achster Konturpunkt
Point $A$ is the annotated centre of the bottleneck ground truth $M_{GT}$. %, which is marked as a green cross in the figure \ref{fig:description_metric}. 
The respective detected points $c^m_{t_1}$ are marked with $B_1$ (inside the ground truth mask  $M_{GT}$) and $B_0$ (outside), and $C_0,C_1$ are indicating the nearest corresponding points on the contour of $M_{GT}$.
The localization error $\epsilon_d$ is defined within $M_{GT}$ between $[0, 1]$ and outside $M_{GT}$ between $(1,\infty]$.
Figure \ref{fig:description_metric} shows that $\epsilon_d$ reaches the value of $0$ if a detected point lies within the smallest assumed isometric contour of $M_{GT}$. 
If the detected point lies on the green contour, $\epsilon_d$ becomes equal to $1$.
%
% The ground truth as well as the evaluation script are publicly available for other work\footnote{\scriptsize{Added upon acceptance of the paper}}.
%
%
%The evaluation of the results, which are shown in figure \ref{fig:results_1}, is done with the considered combination of the localization error $\epsilon_d$ and the accuracy value. 
The results of the evaluation are shown in Figure \ref{fig:results_1} and shows the achieved accuracy in dependence of the localization error $\epsilon_d$.
The higher the accuracy value, the better the events were detected. 
The smaller the localization error, the more accurately the event was detected spatially.

\begin{figure}
\centering
\scriptsize
\begin{tabular}{cc} % entrance_final_result_timeFiltering_4700
   \includegraphics[width=0.21\textwidth]{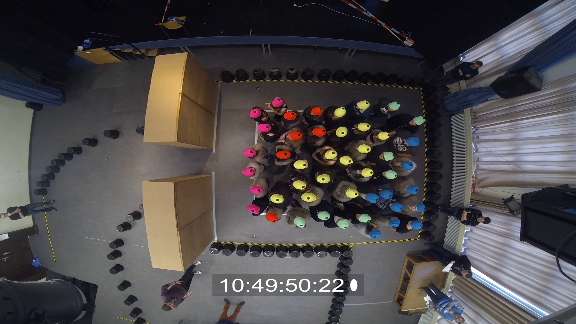} &
  \includegraphics[width=0.21\textwidth]{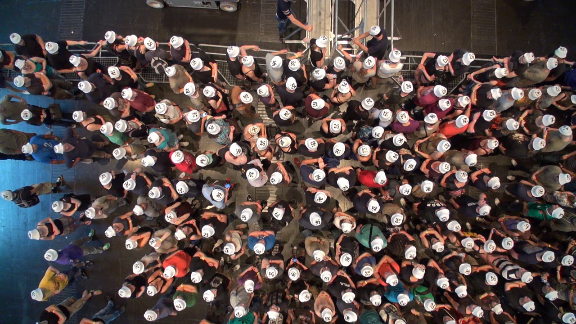} 
  \\  %\vspace{0.05cm}\\
    \includegraphics[width=0.21\textwidth]{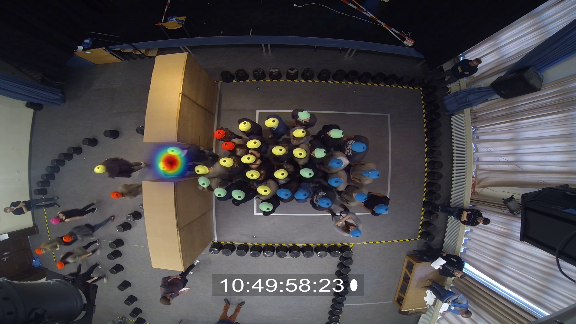} &
  \includegraphics[width=0.21\textwidth]{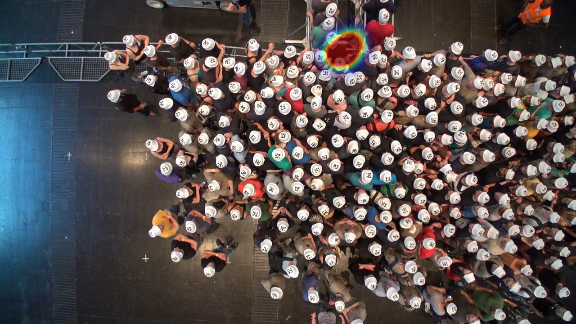}  % scene_04_x4_view_1_final_result_timeFiltering_0285
\end{tabular}
\caption{Detection results of the method for the sequences \textit{Bottleneck and social groups 01\_02} (left) and \textit{Entrance 1, entry without guiding barriers (semicircle setup)} (right). At the beginning of the sequences there is no bottleneck (top).  In the later part of the sequence the crowd starts running respectively the gate is opened.}
\label{fig:results_3}
\end{figure}

The accuracy increases in all results with increasing distance from the optimal point, i.e. with increasing $\epsilon_d$.
The arrangement in Figure \ref{fig:results_1} from left to right shows different parameter tests. 
The Accuracy for a Localization Error value of $\epsilon_d = 1$ with the best settings are listed in Table 1.
%
%Figure \ref{fig:results_1} (top) shows the evaluation results for all eighty sequences.
Figure \ref{fig:results_1} (top) shows the evaluation results for all 80 sequences.
%
%The order within the legends arranges the results after the accuracy for $\epsilon_d = 1$.
%
%The value at this position is given in brackets behind it.
% EB: steht ja schon in der caption
%
The results for the entire test dataset show that the proposed method performs well within the ground truth ($\epsilon_d = 1$) with an accuracy value of $0.7$.
The neighborhood of the optimal point of the bottleneck ($\epsilon_d = 0$) could be reached with an accuracy value of $0.4$.
% EB: Das sollte nicht so stehen bleiben, die 40% kommen eher von den true negatives als von den perfekten treffern
%
Visual results can be seen in Figure \ref{fig:results_3}.
Furthermore, it becomes clear that the results for a small integration time $\tau$ (Figure \ref{fig:results_1} (top, left) are in accordance with the method, because a small $\tau$ also means a lower calculation effort for the calculated path lines. 
Seen over all sequences, the buffer size parameter $\tau_s$ seems to have the smallest effect. 
\begin{figure}
\centering
\scriptsize
\begin{tabular}{cc}
scene04\_x1\_view1 & scene04\_x1\_view2 \\
  \includegraphics[width=0.21\textwidth]{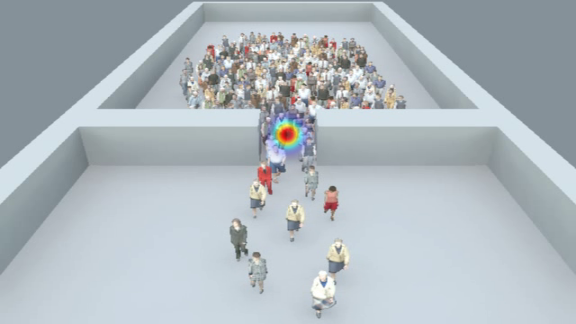} &
  \includegraphics[width=0.21\textwidth]{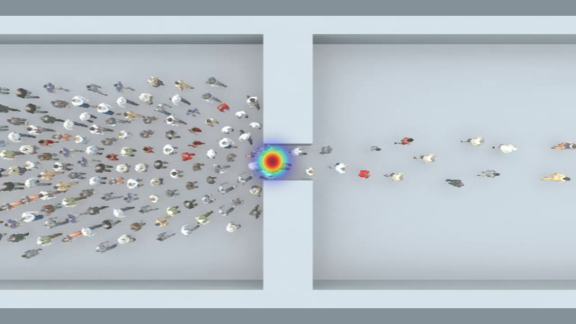} % scene_04_x4_view_1_final_result_timeFiltering_0285
  \\
scene04\_x4\_view1 & scene04\_x4\_view2 \\
  \includegraphics[width=0.21\textwidth]{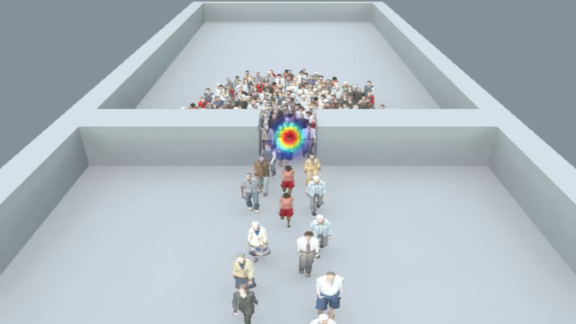} &
  \includegraphics[width=0.21\textwidth]{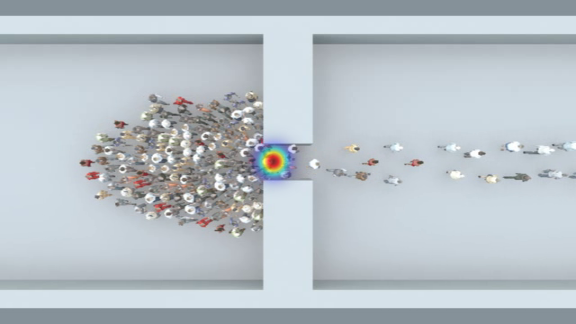}
\end{tabular}
\caption{Detection result of the method for the same time of the escape sequences of AGORASET, at different viewing angles and same parameters.}
\label{fig:results_agoraset}
\end{figure}
This is due to the length of the AGORASET sequences. % TODO ..SOCIAL!!!
The larger the buffer, the more stable the results can be.
However, by averaging the $\tFTLE$ values with the help of the median, the method becomes sluggish, so that the buffer size also presents itself as a limitation of the system.
The result of the partial sequence Entrance\_1 (Figure \ref{fig:results_1}) shows a different effect regarding to the buffer size.
For the largest buffer value, the best result is obtained here.
The sequence is very long and only at the end of the sequence an entry is opened, which represents the bottleneck.
A large buffer has more frames of reference, so that many small movements of the group can be caught, which can lead to errors with smaller buffers.
In the evaluation of the radius $\sigma_r$ it turns out that a smaller value $\sigma_r = 30$ over all sequences achieves the best result.
This is due to the number of detected bottlenecks.
A large radius for the ROI can also enclose unrelated ridges in the validation map.
Certainly there are also sequences in the test dataset which have very large bottlenecks related to the image content.
The filter fails because the ridges in the validation map cannot be included at all.

Figure \ref{fig:results_agoraset} shows the result of the AGORASET escape sequences for the same point in time from two different perspectives. 
The outcome emphasizes that the presented procedure can act independently of the point of view.

  \section{Conclusion}

%  \texttodo{
%   Teile von Einleitung we proposed ... FTLE ... 
%   The evaluation showd that our novel  detection approach for bottlenecks  receives at the current state of development satisfying results but is limited to a fixed size of BN... In future work we want to adjust the ROI depending the defect couples by applying an additional filtering approach. }
  In this paper we presented a novel video-based bottleneck detection method for crowded scenes based on the evaluation of characteristic stowage patterns in crowd-movements. 
  % (FTLE) würde ich hier raus nehmen sonst siehts nach copy paste aus
  We utilized the proposed long-term temporal filtered Finite Time Lyapunov Exponents (FTLE) fields for a global segmentation of the crowd flow, which enables to extract its deformations. 
  Furthermore we showed that high ridges in the FTLE field indicate Lagrangian features that are assumed to be located at bottlenecks. 
  
  % results:
  Ground truth data was generated for 80 tested sequences,  which were evaluated in dependence of the localization error.
%   For the evaluation we have created ground truth data for the eighty sequences we have used, which we make available.
%   %
%   Additionally the Localization Error was presented for the evaluation.
  %  
  The results show that the method can detect bottleneck events spatially and temporally well for both natural and synthetic data. 
  %
  % This is possible because the detection depends less on the actual image content than on the motion fields of the sequence. 
  %
  Our method is independent from camera angle and distortion, but is currently limited in the width of the bottleneck due to a fixed size of the region of interest.   
  % But there are limitations in the width of the bottleneck.
  % future work
  For future work, an adaptive adjustment of the search area is planned, whereby the current restriction will be solved.

\section{Acknowledgements}
%After acceptance of the paper.
%Added upon acceptance.
The  research  leading  to  these  results  has  received  funding BMBF-VIP+ under grant  agreement  number  03VP01940 (SiGroViD).

{\small
\bibliographystyle{ieee}
\bibliography{egbib}
}

%  \rule{0pt}{1pt}\newpage
%  \rule{0pt}{1pt}\newpage
% \rule{0pt}{1pt}\newpage
% \rule{0pt}{1pt}\newpage
% \rule{0pt}{1pt}\newpage

\end{document}